\documentclass[10pt,twocolumn,letterpaper]{article}
\usepackage{cvpr}
\usepackage{amsmath,amssymb,amsthm,bm}
\usepackage{booktabs,multirow,array,tabularx}
\usepackage{graphicx}
\usepackage{xcolor}
\usepackage{colortbl}
\usepackage{xspace}
\usepackage{enumitem}
\usepackage{microtype}
\usepackage{subcaption}
\usepackage{makecell}
\definecolor{cours}{HTML}{EB6834}
\definecolor{cbase}{HTML}{2A78D6}
\definecolor{cgen}{HTML}{6250D6}
\definecolor{cjoint}{HTML}{1BAF7A}
\definecolor{rowgray}{HTML}{F3F2EE}
\newcommand{\method}{RePrime\xspace}
\newcommand{\methodJ}{RePrime-J\xspace}
\newcommand{\fdr}[1]{\ensuremath{\mathrm{FD}_r^{#1}}\xspace}
\newcommand{\sg}{\operatorname{sg}}
\newcommand{\E}{\mathbb{E}}
\newcommand{\R}{\mathbb{R}}
\newcommand{\Lrec}{\mathcal{L}_{\mathrm{rec}}}
\newcommand{\Lalign}{\mathcal{L}_{\mathrm{align}}}
\newcommand{\Lot}{\mathcal{L}_{\mathrm{OT}}}

\newcommand{\suppTitle}{%
  \par\vspace{0.4em}%
  {\centering
   {\large\bfseries Align the Pair, Not the Sample:\\[2pt]
    Representation Alignment for Faster and Better One-Step Generators\par}%
   \vspace{3pt}%
   {\normalsize Supplementary Material\par}%
  \par}%
  \vspace{0.9em}%
}
\newtheorem{proposition}{Proposition}

\setlist[itemize]{leftmargin=*,topsep=2pt,itemsep=1pt,parsep=0pt}
\definecolor{cvprblue}{rgb}{0.21,0.49,0.74}
\usepackage[pagebackref,breaklinks,colorlinks,allcolors=cvprblue]{hyperref}

\title{Align the Pair, Not the Sample:\\Representation Alignment for Faster and Better One-Step Generators}
\author{
Mingxiao Li$^{1}$ \quad Xuantuo Huang$^{1,2}$ \quad Encheng Su$^{1,3}$ \quad Aoran Wang$^{1}$ \\
Han Zhang$^{1}$ \quad Jinouwen Zhang$^{1}$ \quad Shixiang Tang$^{1,4}$ \\[6pt]
{\footnotesize $^1$Shanghai Artificial Intelligence Laboratory \quad $^2$Peking University  \quad
$^3$University of Science and Technology of China \quad $^4$University of Oxford} \\[6pt]
{\small \href{https://mingxiao-li.github.io/RePrime/}{\textcolor{blue}{https://mingxiao-li.github.io/RePrime/}}}
}

\begin{document}
\twocolumn[{%
\renewcommand\twocolumn[1][]{#1}%
\maketitle
\begin{center}
\vspace{-18pt}
\begin{minipage}[c]{0.705\textwidth}
\centering
\includegraphics[width=\linewidth]{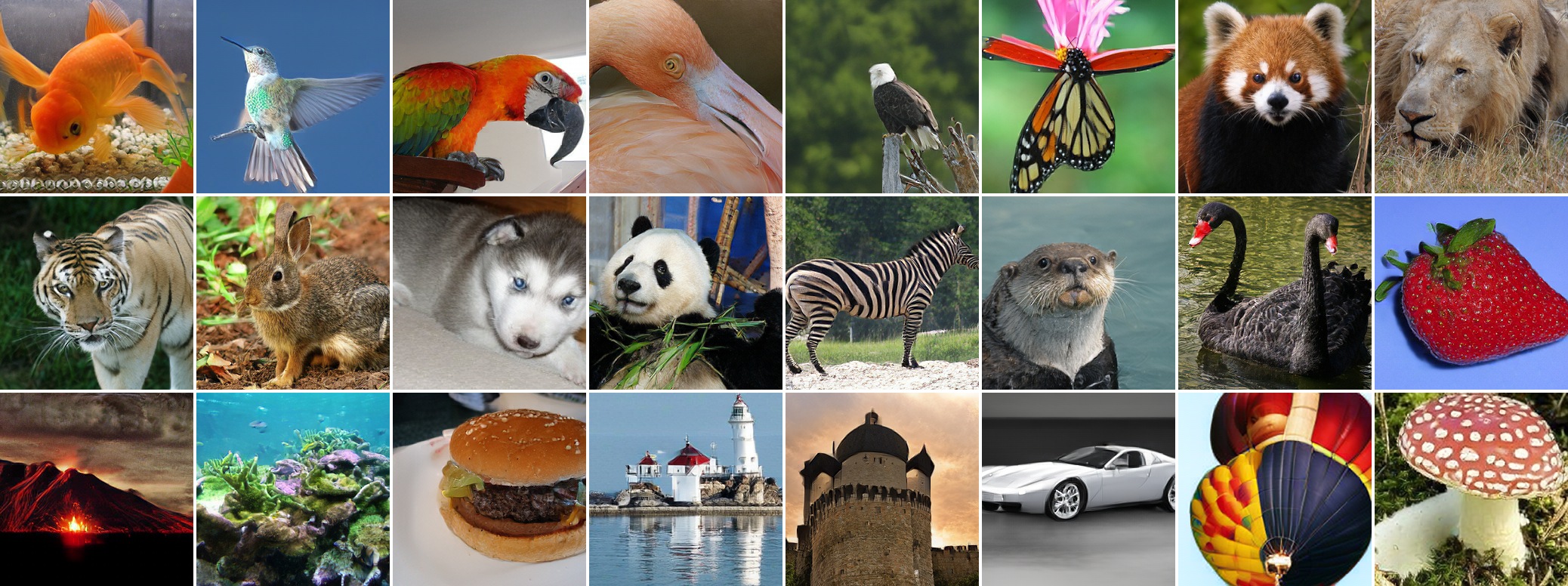}
\end{minipage}\hfill
\begin{minipage}[c]{0.285\textwidth}
\centering
\includegraphics[width=\linewidth]{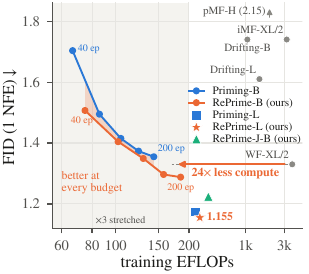}
\end{minipage}
\captionof{figure}{\textbf{Left: one-step ImageNet $256\!\times\!256$ samples from \method-L} (a single network evaluation, no guidance; FID 1.155). \textbf{Right: training compute (teacher included) vs.\ FID} (log axis stretched $3\times$ below 200). \method improves Priming~\cite{priming} at every budget and scale; \method-B beats WF-XL/2 with $24\times$ less compute. Others from~\cite{priming}.}
\label{fig:teaser}
\vspace{6pt}
\end{center}
}]
\begin{abstract}
Representation alignment markedly accelerates diffusion training, but one-step generators raise a question that diffusion does not: \emph{which training path should receive the loss?} A one-step generator sees two kinds of inputs. Corrupted real images come with their clean counterpart, which provides an instance-paired target. Samples generated from pure noise have no corresponding real image, yet this is where the final model operates. We study this question on Priming, which trains one-step generators from scratch by reconstructing corrupted images and then matching feature distributions with optimal transport. Aligning the generative path, the natural choice, collapses recall ($0.45\!\to\!0.15$--$0.25$) in all 17 configurations we test, including five target constructions and loss weights down to $0.05$. Aligning the paired reconstruction path with a frozen DINOv2 encoder instead helps in every setting we test. A short analysis explains this asymmetry: when the target carries no instance information about the input, the optimal aligned feature is constant within each class, so the loss pulls samples toward class prototypes. A control confirms this: aligning the paired path to a random same-class image instead of the input's own image removes most of the gain. Inside the generator, paired-path alignment creates an interface block through which post-training routes noise, and the image semantics carried through it predict sample quality across 16 models. The resulting recipe, \method, accelerates \emph{both} training stages on ImageNet $256\!\times\!256$: it matches the 160-epoch baseline with $2\times$ fewer priming epochs, surpasses its final quality with $4\times$ fewer post-training updates, and lowers FID to 1.30 (B) and 1.155 (L) at the standard budget. A joint single-stage variant trains native one-step generators and reaches FID 1.22 with 115M parameters.
\end{abstract}

\section{Introduction}
\label{sec:intro}
Diffusion models now drive high-fidelity image and video synthesis~\cite{dit,animatemotion,zigzagstory,li2025personalized}, but every sample costs tens to hundreds of network evaluations, which makes generation slow and lets errors compound along the sampling trajectory~\cite{ning2024exposure,li2024shifted}. One-step generators remove this cost by mapping noise to an image in a single forward pass. They can be distilled from a diffusion teacher~\cite{dmd,dmd2,consistency} or trained from scratch with flow-map~\cite{shortcut,meanflow,imf,pmf} and distributional objectives~\cite{drifting,wflow,semantictransport}. Priming~\cite{priming} made the from-scratch route remarkably efficient: a pixel-space transformer is first \emph{primed} by reconstructing heavily corrupted images, $x_s=(1-s)x_0+s\epsilon$ with $s=0.8$, and then post-trained on pure noise by matching its samples to real images with entropic optimal transport (OT) in frozen feature spaces, reaching FID 1.19 on ImageNet with a tenth of the compute of prior one-step generators. Training nevertheless remains costly, with 160 priming epochs and OT updates that each cost about five priming updates. Diffusion training faced the same problem and found a powerful remedy: denoising objectives learn semantics slowly, and REPA~\cite{repa} accelerates them by an order of magnitude by aligning intermediate features with a frozen self-supervised encoder~\cite{repa,irepa,haste,repae}. This raises a natural question: \emph{can representation alignment make one-step generators faster to train and better?}

\begin{figure*}[t]
\centering
\includegraphics[width=\textwidth]{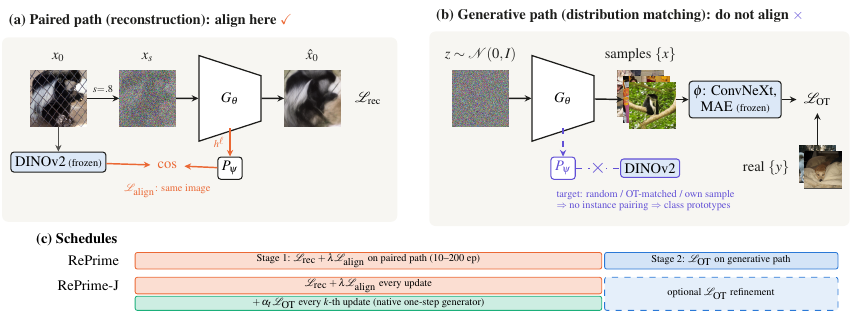}
\vspace{-14pt}
\caption{\textbf{Where to align a one-step generator.} (a) On the \emph{paired path}, the generator reconstructs a corrupted real image $x_s$, and an intermediate feature $h^\ell$ is aligned, through a light projector $P_\psi$, with frozen DINOv2 features of the \emph{same} clean image $x_0$. (b) On the \emph{generative path}, the same weights map noise to samples that are matched to real images by OT in frozen ConvNeXt/MAE feature spaces. A sample has no corresponding real image, so every alignment target we construct lacks instance correspondence and collapses coverage (\cref{sec:placement}). (c) \method aligns only during priming; \methodJ trains both paths jointly.}
\label{fig:method}
\vspace{-6pt}
\end{figure*}

Answering it is harder than transplanting REPA. In diffusion training, every input is a corrupted real image, so the alignment target, the encoder's view of the clean image, is always defined. A one-step generator instead learns from \emph{two} kinds of inputs (\cref{fig:method}). On the \emph{paired path}, it reconstructs a corrupted real image whose clean counterpart is known; on the \emph{generative path}, it maps pure noise to samples that have \emph{no} corresponding real image. Neither path is an obvious home for the loss. The generative path is where the final model operates, and one-step objectives already compare samples with real images there in frozen feature spaces~\cite{priming,drifting,semantictransport,fdloss,sf2}. Yet a sample has no partner, so its target must be constructed, \eg, from a random same-class image, from the image that the OT plan matches to it, or from the sample itself, and such a target may say nothing about \emph{this} sample, pulling a class together and eroding the diversity that distribution matching works to preserve. The paired path has a well-defined target but exists only in training; as with the training--inference gap behind exposure bias~\cite{ning2024exposure,li2024shifted}, semantics learned on corrupted images need not reach the noise inputs from which the model generates. Moreover, since the OT stage already matches samples in frozen feature spaces, a teacher may merely duplicate what post-training sees, and since a frozen teacher makes every update more expensive, any gain must survive a comparison at matched total compute.

To address the challenges above, we first ask which path can support alignment at all through an \emph{alignability analysis} (\cref{sec:theory}). When the target carries no instance information about the input, the optimal aligned feature is constant within each class, so the loss can only pull all samples of a class toward one prototype; when the input reveals its target, as $x_s$ still reveals $x_0$, the optimum must encode instance-specific semantics. Guided by this analysis, \method (\emph{Representation-aligned Priming}) applies \emph{paired-path alignment} during priming only: a light projector maps the patch tokens of an intermediate block to the frozen encoder's features of the same clean image and is discarded afterwards, so inference is unchanged. As encoder we use a \emph{cross-space semantic teacher}, DINOv2, which lies outside the ConvNeXt and MAE spaces of the OT stage and thus supplies semantics that post-training cannot obtain on its own. An \emph{interface analysis} then tests whether these semantics reach the generative path by locating the block where features of noise and of corrupted inputs meet and measuring how much image semantics noise routes through it (\cref{sec:mechanism}). Finally, \methodJ interleaves OT updates on noise with aligned reconstruction, yielding a \emph{native} one-step generator without a separate post-training stage while still aligning the paired path only.

The experiments reveal a sharp asymmetry. On the generative path, alignment fails in \emph{all 17} configurations we test, spanning five target constructions, two feature spaces, two loss forms, and loss weights down to 0.05: recall falls from 0.45 to 0.15--0.25 at maintained or higher precision, the prototype attraction that the analysis predicts. On the paired path, the same kind of loss helps in all 20 comparisons; a random same-class target removes most of the gain, and the semantics routed through the interface predict sample quality across 16 models. On ImageNet $256\!\times\!256$, \method accelerates \emph{both} stages, matching the 160-epoch baseline with $2\times$ fewer priming epochs and surpassing its final quality with $4\times$ fewer OT updates. It is better at all seven priming budgets and at matched compute (\cref{fig:teaser}, right), and lowers FID from 1.37 to 1.30 (B) and from 1.17 to 1.155 (L), with 9--13\% lower \fdr{5}; \methodJ reaches FID 1.22 with 115M parameters.

The loss itself is REPA's; our contribution is where it must act, why, and evidence from controls and from inside the generator. We establish a \emph{placement principle}, \emph{align the pair, not the sample}, that holds in two-stage and single-stage training; we turn it into \method, a drop-in change to priming that reaches the best one-step FID (1.155) among the methods compared in~\cite{priming}; and we show that a semantic teacher outside the OT feature spaces is essential, that ease of alignment is \emph{anti}-correlated with benefit, that mid-depth targets work best, and that, unlike in diffusion, stopping alignment early does not help.

\section{Related Work}
\label{sec:related}
\paragraph{One-step generation.}
Distillation methods compress a pretrained diffusion model into one or few steps~\cite{consistency,dmd,dmd2}. Training from scratch has been revisited through flow maps and average velocities~\cite{shortcut,imm,meanflow,imf,pmf}, and through distributional objectives that match generated and real samples directly~\cite{gmmn,genevay2018,drifting,wflow,semantictransport,fdloss}. Priming~\cite{priming} shows that a cheap reconstruction stage at a fixed noise level makes the subsequent distributional stage, entropic OT~\cite{sinkhorn,sinkhorndivergences} in frozen ConvNeXt and MAE feature spaces, far more efficient. Like recent pixel- and frequency-space diffusion models~\cite{jit,dctdiff}, it needs no learned tokenizer. We ask where in this recipe an auxiliary feature loss should act.

\paragraph{Representation alignment for generation.}
REPA~\cite{repa} aligns intermediate diffusion-transformer features of a noisy input with a frozen encoder's features of the clean image. Follow-ups study which properties of the target matter, finding spatial structure important~\cite{irepa}, when to stop aligning~\cite{haste}, end-to-end training with the tokenizer~\cite{repae}, how to shape the latent space itself for diffusability~\cite{vavae,rae,spectrummatching}, and encoder-free regularizers~\cite{dispersive}. In all of these, the aligned input is a corrupted real image with an observed clean counterpart. A one-step generator also trains on noise-only inputs without such a counterpart. This raises a question that does not arise in diffusion: \emph{which training path should be aligned?}

\paragraph{Representation objectives in distribution matching.}
SF$^2$~\cite{sf2} adds a predictive representation loss to DMD for few-step video distillation, using views from a shared generation trajectory and an EMA teacher. EASD~\cite{easd} assigns score and adversarial objectives to different noise regimes. We answer it for from-scratch one-step training with feature-space OT: instance-anchored alignment on the reconstruction path consistently helps, whereas unanchored targets on the generative path consistently trade coverage for typicality. Semantic Transport~\cite{semantictransport} shows that the \emph{matching} space matters for OT-based one-step generation; we find a complementary effect for the \emph{alignment} space, which should differ from the matching spaces.

\section{Method}
\label{sec:method}

\subsection{Preliminaries: priming}
\label{sec:prelim}
Let $G_\theta(\cdot,c)$ be a class-conditional pixel-space transformer that maps a $256\!\times\!256$ input to an image of the same size in one forward pass~\cite{priming}. Priming trains $G_\theta$ in two stages.

\noindent\textbf{Stage 1 (priming)} reconstructs corrupted real images at a fixed noise level,
\begin{equation}
x_s=(1-s)\,x_0+s\,\epsilon,\quad \epsilon\sim\mathcal N(0,I),\quad s=0.8,
\label{eq:corrupt}
\end{equation}
by minimizing $\Lrec(G_\theta(x_s,c),x_0)$, an adaptively normalized pixel error plus LPIPS~\cite{lpips} and ConvNeXt~\cite{convnext} perceptual terms.

\noindent\textbf{Stage 2 (post-training)} feeds pure noise $z\sim\mathcal N(0,I)$ and matches, class by class, the distribution of samples $X=G_\theta(z,c)$ to real images $Y$ in frozen feature spaces $\phi_j$ (ConvNeXt and ResNet-MAE):
\begin{equation}
\Lot=\sum_j w_j\Big[D_j\big(\phi_j(X),\phi_j(Y)\big)-D_j\big(\phi_j(X),\sg[\phi_j(X')]\big)\Big],
\label{eq:ot}
\end{equation}
where $D_j$ is an entropic OT cost computed with Sinkhorn iterations~\cite{sinkhorn,sinkhorndivergences}, $X'$ is a second, stop-gradient batch of samples, and $w_j$ are the calibration weights of~\cite{priming}. At inference, one evaluation $G_\theta(z,c)$ generates an image.

\subsection{Two paths, one alignment loss}
\label{sec:paths}
Representation alignment~\cite{repa} adds a loss on the patch features $h^\ell\in\R^{N\times D}$ after block $\ell$:
\begin{equation}
\Lalign=\frac1N\sum_{p=1}^{N}\Big(1-\cos\big(P_\psi(h^\ell_p),\,T_p\big)\Big),
\label{eq:align}
\end{equation}
where $P_\psi$ is a small trainable projector and $T\in\R^{N\times D'}$ is a target map from a frozen encoder $E$. During priming the input is $x_s$, and the natural target is the encoder's view of the \emph{same} image, $T=E(x_0)$; we call this the \emph{paired path}. During post-training the input is $z$, and no real image corresponds to the sample, so $T$ must be constructed. We consider every construction we found plausible: the features of (i)~a random real image of the same class, (ii)~a token pool of several same-class images, (iii)~the real image to which the OT plan of \cref{eq:ot} assigns the most mass, (iv)~the generated sample itself, and (v)~an EMA teacher's sample for the same noise. We call this the \emph{generative path}.

\subsection{Why the path matters: alignability}
\label{sec:theory}
Consider one target token. Let $X$ denote the generator input and $c$ its class, let $u(X)$ be the unit-normalized prediction, and let $T$ be the unit-normalized target. The alignment risk is $\mathcal R(u)=\E[1-u(X)^\top T]$.

\begin{proposition}[Alignability]
\label{prop:align}
Let $\mu(X)=\E[T\mid X]$ and $m_c=\E[T\mid c]$. Then
\begin{enumerate}[label=(\roman*),leftmargin=*,topsep=1pt]
\item $\mathcal R(u)\ge 1-\E\|\mu(X)\|$, with equality iff $u(X)=\mu(X)/\|\mu(X)\|$ wherever $\mu(X)\ne0$;
\item if $T$ is independent of $X$ given $c$, the optimal prediction is \emph{constant within each class}, $u^\star(X)=m_c/\|m_c\|$;
\item in general, $\E\|\mu(X)\|\ge\E_c\|m_c\|$. The gap $\mathcal A=\E\|\mu(X)\|-\E_c\|m_c\|$ is zero iff the optimal prediction is constant within each class.
\end{enumerate}
\end{proposition}
\begin{proof}[Proof sketch]
$\mathcal R(u)=1-\E[u(X)^\top\mu(X)]\ge1-\E\|\mu(X)\|$ by Cauchy--Schwarz, since $\|u\|=1$. Under conditional independence, $\mu(X)=m_c$. Part (iii) follows from Jensen's inequality, $\E[\|\mu(X)\|\mid c]\ge\|m_c\|$, with equality iff the $\mu(X)$ within a class are positively collinear. A full proof is in the supplement.
\end{proof}

\noindent\textbf{Implications.} The proposition is elementary; its value is a falsifiable prediction (tested in \cref{sec:placement}). The excess $\mathcal A$ measures how much instance information about the target the input carries. On the paired path, $x_s=0.2\,x_0+0.8\,\epsilon$ still reveals $x_0$: after 160 epochs of priming, the generator's reconstruction of $x_s$ is closer in DINOv2 space to its own clean image than to its class prototype (cosine 0.81 vs.\ 0.73; supplement). Hence $\mathcal A>0$, and alignment rewards features that encode the semantics of \emph{this} image. On the generative path with random or pooled targets, (ii) applies exactly: the optimum is a class-constant direction, and the gradient pulls the features of \emph{every} noise input of class $c$ toward the same prototype. OT-matched targets depend on the sample only through a plan computed in ConvNeXt/MAE space; in DINOv2 space the matched image is barely closer to the sample than a random same-class one (cosine 0.221 vs.\ 0.213; supplement). Self- and EMA-targets move with the generator and fall outside the proposition, yet show the same symptom: higher precision, collapsed recall (\cref{sec:placement}).

\subsection{\method: align the pair, not the sample}
\label{sec:reprime}
\method applies \cref{eq:align} on the paired path only:
\begin{equation}
\mathcal L_{\text{Stage 1}}=\Lrec\big(G_\theta(x_s,c),x_0\big)+\lambda\,\Lalign\big(h^\ell(x_s,c),\,E(x_0)\big),
\label{eq:reprime}
\end{equation}
and leaves Stage 2 (\cref{eq:ot}) unchanged. $E$ is DINOv2-L/14~\cite{dinov2}, applied to $x_0$ resized to $224^2$. Its $16\!\times\!16$ patch tokens align one-to-one with the generator's patch grid (class tokens excluded). $P_\psi$ is a two-layer GELU MLP (width 1024) trained with AdamW (learning rate $10^{-4}$) and discarded after priming, so inference is unchanged. We use $\lambda=0.5$, keep alignment on throughout priming, and align block $\ell=5$ of 16 for B (block 8 is better; \cref{sec:ablation}) and block 10 of 32 for L.

\noindent\textbf{\methodJ: joint priming.} The two-stage recipe learns semantics on one path and only later teaches the other path to use them. If, as we show in \cref{sec:mechanism}, post-training mainly routes noise through features formed during priming, the two paths can also be trained at once, so that noise learns to use these features while they form. \methodJ interleaves both objectives; at outer update $t$,
\begin{equation}
\mathcal L_t=\mathcal L_{\text{Stage 1}}+\alpha_t\,\mathbb 1[t\bmod k=0]\,\Lot,
\label{eq:joint}
\end{equation}
with $k=2$ and $\alpha_t$ ramped linearly from 0 to 1 over the first 25k updates. The two losses use separate forward passes and one optimizer step. Although the generative path is now trained throughout, alignment still acts only on the paired path: \cref{prop:align} applies regardless of the schedule, and the generative path receives the teacher's semantics only through the shared weights. The resulting \emph{native} model needs no post-training but can be refined with Stage 2.

\noindent\textbf{Cost.} The teacher adds one DINOv2-L forward pass per training image (162 GFLOPs), \ie, $+43\%$ FLOPs per B priming update and $+19\%$ for L; the projector is negligible. Because Stage 2 dominates the compute at low budgets, this amounts to $+4$--$29\%$ of total training compute at B scale and $+14\%$ at L scale (\cref{tab:main}). An OT update costs about $4.9\times$ a priming update, so \methodJ with $k=2$ costs about $3.9\times$ a Priming update. We report all costs in EFLOPs, following~\cite{priming}.

\section{Experiments}
\label{sec:exp}

\subsection{Setup}
\label{sec:setup}
We follow Priming~\cite{priming} on class-conditional ImageNet $256\!\times\!256$~\cite{imagenet} with its pixel-space B (115M parameters) and L (405M) generators and its recipe: 160 priming epochs (200,160 updates) followed by 25k post-training updates of \cref{eq:ot} (hyper-parameters in the supplement). \method differs from the baselines only by the loss in \cref{eq:reprime}. Baselines post-train the released Priming-B/L initializations, which reproduce the reported FIDs; budget sweeps retrain the official recipe.

\noindent\textbf{Metrics.} We generate 50k samples with one network evaluation and no guidance. We report FID~\cite{fid} and \fdr{5} under the official protocol of~\cite{priming}. \fdr{5} averages normalized Fr\'echet distances in Inception, DINOv2, ViT-MAE, SigLIP, and CLIP spaces~\cite{fdloss}; \fdr{6} adds ConvNeXt and comes with precision and recall~\cite{precisionrecall} from the training-time evaluator. Main results average three sampling seeds; \fdr{6}, precision, and recall use EMA weights.

\subsection{Where to align: paired vs.\ generative path}
\label{sec:placement}
\noindent\textbf{Protocol.} Starting from the released Priming-B initialization, we post-train for 2.5k updates with \cref{eq:ot} plus an alignment loss on block 8, switched on after 250 updates. We test the five targets of \cref{sec:paths} in DINOv2 and OT ConvNeXt space with a max-over-tokens cosine (10 configurations), and a tokenwise cosine with random and OT-matched targets, $\lambda\in\{1,0.25,0.1,0.05\}$, and windows ending after 750 updates (7 configurations). If generative-path alignment were merely poorly tuned, some of these configurations should help; if \cref{prop:align} is right, all of them should pull samples toward class prototypes, raising precision and lowering recall. As a paired-path counterpart under the \emph{same} protocol, we post-train a \method-B initialization with plain \cref{eq:ot}.

\begin{figure}[t]
\centering
\includegraphics[width=\linewidth]{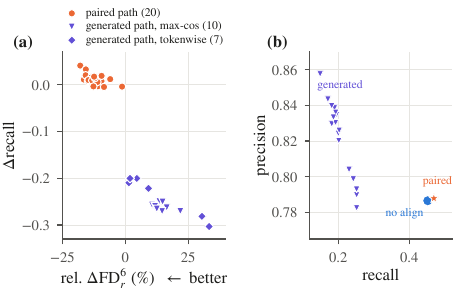}
\vspace{-14pt}
\caption{\textbf{Placement decides the sign of alignment.} (a) Change in \fdr{6} and recall vs.\ the matching no-alignment baseline: 20 paired-path comparisons and 17 generative-path configurations. (b) Precision--recall under the same 2.5k-update protocol.}
\label{fig:placement}
\vspace{-2pt}
\end{figure}

\begin{figure}[t]
\centering
\includegraphics[width=\linewidth]{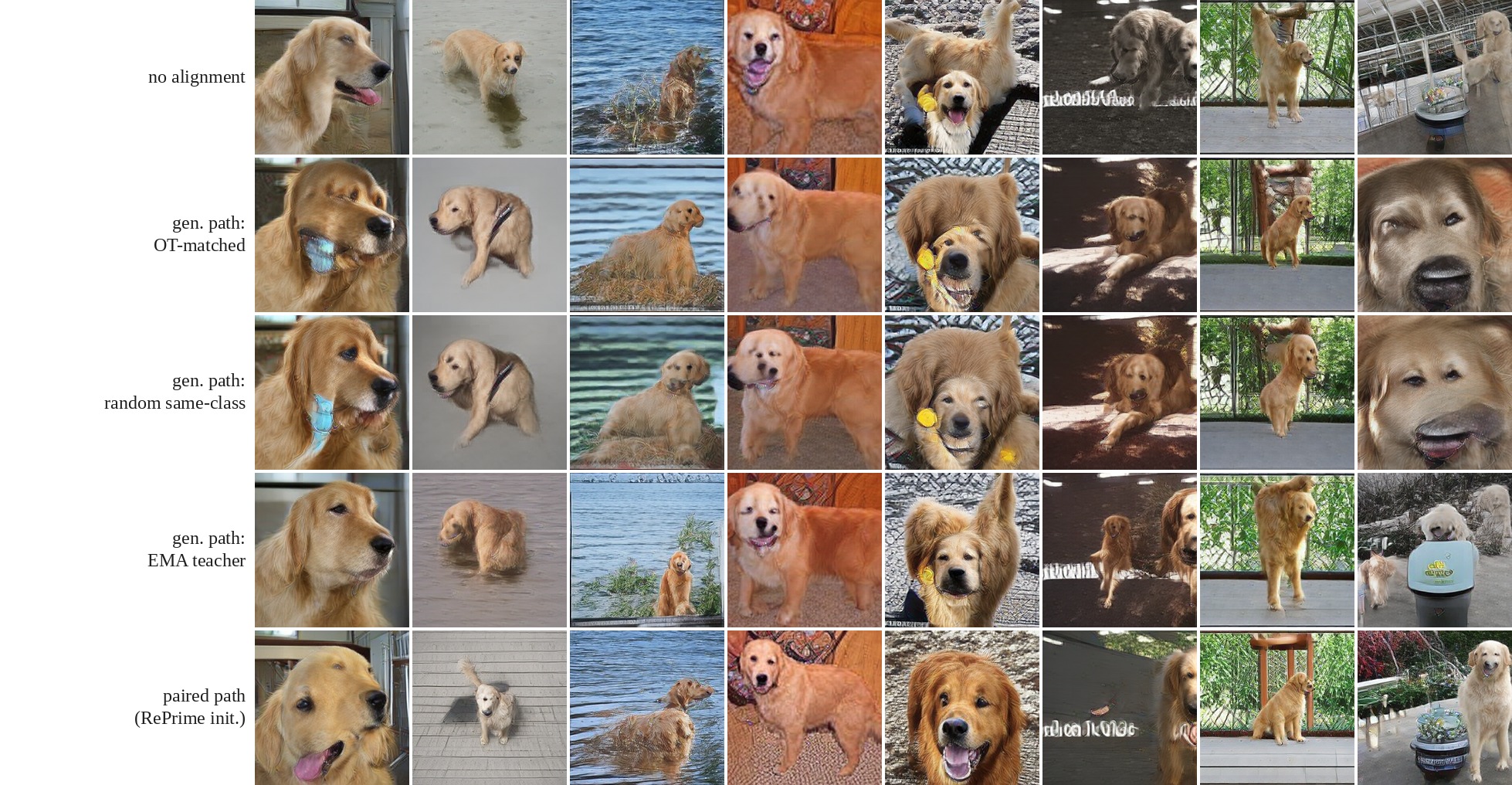}
\vspace{-14pt}
\caption{\textbf{Prototype attraction} (class 207, the \emph{same} 8 noise vectors in every row, 2.5k post-training updates). Rows 1--4: Priming-B initialization with no alignment and with generative-path alignment to OT-matched, random same-class, and EMA-teacher targets. Row 5: \method initialization. OT-matched and random rows are nearly identical, with close-up, centered dogs on simpler backgrounds (\eg, columns 2, 5, 8). More classes in the supplement.}
\label{fig:collapse}
\vspace{-4pt}
\end{figure}

\noindent\textbf{Generative-path alignment trades coverage for typicality.} All 17 configurations lower recall from 0.452 to 0.149--0.252 and worsen \fdr{6} (14.71$\to$14.92--19.62), while precision stays at or above the baseline in 16 of them (\cref{fig:placement}b). Rising precision with falling recall means that samples concentrate on typical modes of the class rather than becoming individually worse: the prototype attraction predicted by \cref{prop:align}. The failure is not a matter of dose: at $\lambda{=}0.05$ recall still falls to 0.252. It is not transient: switching alignment off after 750 updates and continuing pure OT for 1750 updates leaves recall at 0.242, and 2.5k further pure-OT updates after $\lambda{=}0.05$ alignment raise it only from 0.252 to 0.290, so OT undoes only a small part of the damage. And it is not a poorly chosen target space: in the OT ConvNeXt space the projector matches its targets almost perfectly (cosine 0.998), yet recall falls to 0.18--0.20, so a well-fit target does not help if it says nothing about the sample. \cref{fig:collapse} shows the effect; OT-matched targets act like random ones because, seen by DINOv2, the plan picks partners almost at random within the class (\cref{sec:theory}).

\noindent\textbf{Paired-path alignment improves fidelity and coverage.} Under the same protocol, a \method initialization lowers \fdr{6} from 14.71 to 12.89 and \emph{raises} recall from 0.452 to 0.469 at unchanged precision. Across all 20 paired-path comparisons in this paper (budgets, scales, depths, teachers, post-training lengths, joint training), \fdr{6} improves by 1--18\% and recall never drops by more than 0.005 (\cref{fig:placement}a). The same loss thus collapses coverage on one path and improves it on the other; placement, not the loss, decides the outcome.

\noindent\textbf{Instance correspondence drives the gain.} The comparison above changes both path and target, so we test \cref{prop:align} on the paired path alone, priming for 10 epochs with targets that break correspondence (a random same-class image, or the own image with permuted tokens) and post-training for 5k updates. If alignment helped as a generic regularizer or by injecting class-level semantics, random targets should recover most of the gain. They do not: \fdr{6} is 22.10 without alignment, 21.62 and 21.47 in two random-target runs, 19.94 with permuted tokens, and 18.10 with the own image, \ie, random targets recover only 12--16\% of the gain and permuted tokens 54\%. Recall follows the same order (0.486 to 0.514). The ranking tracks the instance information in the target: a random image carries none, permuted tokens keep the image's content but lose its layout, and the own image keeps both. Alignment thus helps only when the target describes the specific input.

\subsection{Why paired-path alignment helps}
\label{sec:mechanism}

\begin{figure}[t]
\centering
\includegraphics[width=\linewidth]{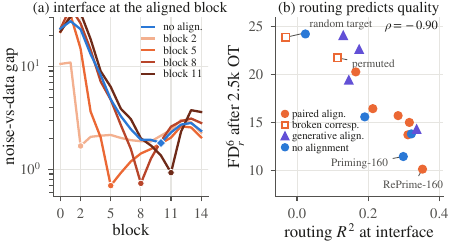}
\vspace{-14pt}
\caption{\textbf{Mechanism} (2.5k OT updates, online weights). (a) Noise-vs-data feature gap by depth for 40-epoch \method aligned at different blocks (dots: aligned block). (b) Semantic routing at the interface vs.\ \fdr{6} for 16 models.}
\label{fig:mechanism}
\vspace{-4pt}
\end{figure}

Alignment acts only on corrupted real images, yet the gain appears when generating from noise. Since post-training moves the primed weights by only 2--4\% (supplement), what it exploits must exist after priming. We thus ask where noise and real inputs meet inside the generator and what passes through that point, feeding 1,024 images of 16 classes as data inputs $x_s$ or as noise $z$ with the same labels and comparing mean-pooled block features.

\noindent\textbf{Alignment creates an interface.} The noise-vs-data gap is the Fr\'echet distance between the two feature sets over the trace of the data covariance. After 2.5k OT updates, the gap is smallest exactly at the aligned block for all four alignment depths (\cref{fig:mechanism}a) and 2.6$\times$ smaller than the minimum without alignment (0.70 vs.\ 1.81). Because the minimum moves with the aligned block, the interface is created by alignment rather than being a fixed property of the architecture. It already exists before post-training (gap 4.5 vs.\ 28.4 at block 5 after 160 epochs), so post-training finds it rather than builds it.

\noindent\textbf{What matters is the semantics routed through it.} A small gap alone does not explain quality: generative-path alignment closes it too (0.47--0.61), largely by squeezing the noise features of a class together (within-class spread 0.13 vs.\ 0.41), and across 16 models the gap is uncorrelated with \fdr{6} (Spearman $\rho{=}0.00$). We therefore measure \emph{semantic routing}: a ridge probe fit on data-mode features at the interface to predict DINOv2 features of the clean image is applied to noise-mode features and scored against DINOv2 features of the generated image. A high $R^2$ means that noise uses the interface like real inputs do. Routing tracks \fdr{6} across all 16 models ($\rho{=}{-0.90}$; \cref{fig:mechanism}b). The correspondence control separates the two quantities: random targets also form an interface at block 5 (gap 1.57), but its features predict held-out CLIP embeddings no better than without alignment ($R^2$ 0.361 vs.\ 0.360; own image 0.413), and noise routes no semantics through it ($-0.03$ vs.\ 0.17). This completes the explanation of the placement principle: instance-paired targets write image-specific semantics into the interface and post-training routes noise through it, whereas unpaired targets create an interface with nothing useful to route.

\subsection{Main results}
\label{sec:main}

\begin{table}[t]
\centering
\footnotesize
\setlength{\tabcolsep}{3.1pt}
\begin{tabular}{@{}l c r c c c c@{}}
\toprule
Method & Ep. & EFLOPs & FID$\downarrow$ & \fdr{5}$\downarrow$ & Prec.$\uparrow$ & Rec.$\uparrow$\\
\midrule
\multicolumn{7}{@{}l}{\textit{Reported in~\cite{priming} (1 NFE)}}\\
pMF-H~\cite{pmf} & -- & 1968.5 & 2.15 & 8.14 & -- & --\\
iMF-XL/2~\cite{imf} & -- & 3201.1 & 1.74 & 9.75 & -- & --\\
Drifting-L~\cite{drifting} & 1280 & 1476.4 & 1.61 & 12.59 & .815 & .602\\
WF-XL/2~\cite{wflow} & -- & 3744.4 & 1.33 & 11.07 & .762 & .674\\
Priming-B~\cite{priming} & 160 & 124.2 & 1.38 & 11.23 & .783 & .624\\
Priming-L~\cite{priming} & 160 & 239.0 & 1.19 & 7.10 & .781 & .649\\
\midrule
\multicolumn{7}{@{}l}{\textit{B, 115M parameters (our runs, identical pipeline)}}\\
Priming-B & 160 & 124.2 & 1.373 & 11.27 & .784 & .620\\
Priming-B & 200 & 143.4 & 1.353 & 10.91 & .784 & .625\\
\rowcolor{rowgray}\method-B & 120 & 129.8 & 1.349 & 10.76 & .784 & .627\\
\rowcolor{rowgray}\method-B & 160 & 157.3 & 1.303 & 10.20 & .781 & .639\\
\rowcolor{rowgray}\method-B & 200 & 184.8 & 1.283 & 9.66 & .784 & .631\\
\rowcolor{rowgray}\method-B, block 8 & 160 & 157.3 & 1.268 & 9.82 & .781 & .641\\
\rowcolor{rowgray}\method-B, block 8 & 200 & 184.8 & 1.266 & 9.66 & .782 & .635\\
\rowcolor{rowgray}\methodJ-B$^\ast$ & 160 & 346.8 & \textbf{1.224} & 10.60 & .779 & .640\\
\midrule
\multicolumn{7}{@{}l}{\textit{L, 405M parameters (our runs, identical pipeline)}}\\
Priming-L & 160 & 239.0 & 1.174 & 7.41 & .779 & .648\\
\rowcolor{rowgray}\method-L & 160 & 272.1 & \textbf{1.155} & \textbf{6.43} & .781 & .644\\
\bottomrule
\end{tabular}
\caption{\textbf{ImageNet $256\!\times\!256$, one network evaluation, no guidance.} EFLOPs count both stages and, for \method, the frozen teacher. ``Ep.'': priming epochs; all models are post-trained for 25k updates. FID/\fdr{5} average three sampling seeds (one for the 120-epoch row; std.\ ${\le}0.022$ FID, ${\le}0.035$ \fdr{5}). \method-B aligns block 5 unless noted. $^\ast$Joint training (\cref{sec:joint}), 200k joint updates followed by 25k refinement updates.}
\label{tab:main}
\vspace{-4pt}
\end{table}

\cref{tab:main} compares \method with Priming under an identical pipeline. At B scale and 160 epochs, \method lowers FID from 1.373 to 1.303 and \fdr{5} by 9.4\% (11.27$\to$10.20) and raises recall from 0.620 to 0.639 at the same precision; as on the paired path in \cref{sec:placement}, alignment improves samples without concentrating them. \method-L reaches FID 1.155, the best one-step FID among the methods reported in~\cite{priming}, below Priming-L (1.19) and WF-XL/2 (1.33), which uses $14\times$ more compute; even \method-B surpasses WF-XL/2 with $24\times$ less compute (\cref{fig:teaser}). The FID gain at L (0.019) is within seed variation, but the semantic metrics show that the benefit does not shrink with scale: \fdr{5} drops by 13.3\% (7.41$\to$6.43), and the relative \fdr{6} gain grows from $-9.5\%$ at B to $-12.7\%$ at L.

\noindent\textbf{The gains exceed run-to-run variation.} A second 160-epoch priming run per recipe agrees with the first to within 0.21 \fdr{6} and 0.006 FID, whereas the recipes differ by 0.9--1.2 \fdr{6} and 0.07 FID (supplement), four to six times the training noise.

\begin{figure*}[t]
\centering
\includegraphics[width=\textwidth]{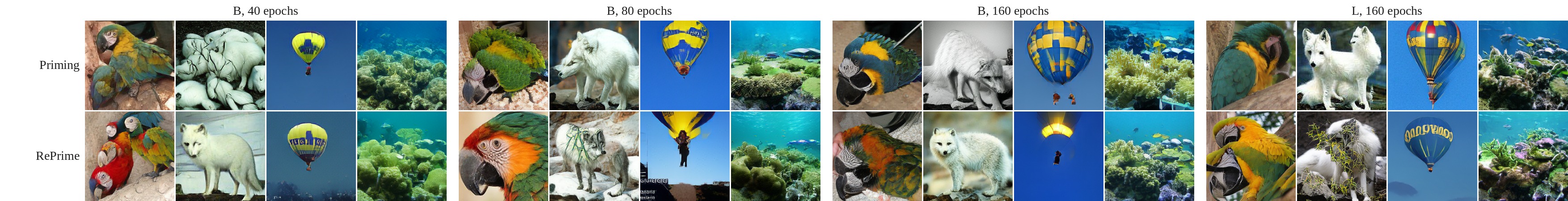}
\vspace{-14pt}
\caption{\textbf{Same noise, different initialization.} Uncurated samples from Priming (top) and \method (bottom) after 40, 80, and 160 epochs of priming followed by 25k post-training updates (B), and at L scale (right). Each column uses the same class and noise.}
\label{fig:qualitative}
\vspace{-6pt}
\end{figure*}

\noindent\textbf{The gains are not a teacher artifact.} DINOv2, our teacher, is one of the five \fdr{5} spaces, so the gain might only reflect agreement with the teacher. Without the DINOv2 term, \fdr{5} still drops by 8.6\% at B (12.46$\to$11.38) and 12.7\% at L (8.19$\to$7.14); every non-teacher space improves at L, and all but CLIP at B. \method thus improves samples as judged by encoders it never saw, and from identical noise it yields more coherent shapes and layouts, especially at low budgets (\cref{fig:qualitative}).

\subsection{Faster training}
\label{sec:speed}

\begin{figure*}[t]
\centering
\includegraphics[width=\textwidth]{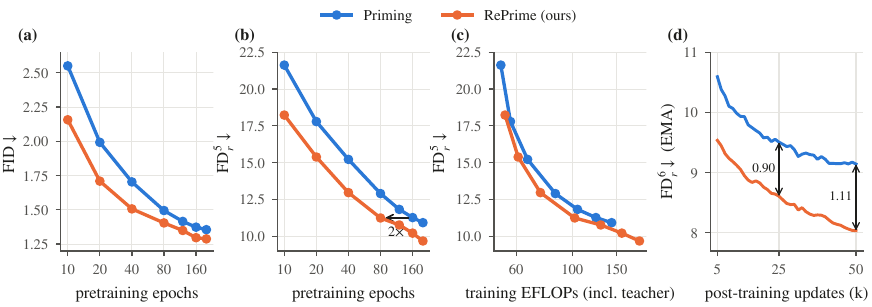}
\vspace{-14pt}
\caption{\textbf{\method accelerates both stages.} (a,b) FID and \fdr{5} after 25k post-training updates vs.\ priming epochs. (c) \fdr{5} vs.\ total training compute, teacher included. (d) Post-training curves from the 160-epoch initializations.}
\label{fig:budget}
\vspace{-6pt}
\end{figure*}

\noindent\textbf{Fewer priming epochs.} \cref{fig:budget}(a,b) varies the priming budget from 10 to 200 epochs, each followed by 25k post-training updates. \method is better at all seven budgets in FID, \fdr{5}, \fdr{6}, and recall. With 80 epochs it matches the 160-epoch baseline (\fdr{5} 11.24 vs.\ 11.27; \fdr{6} 9.43 vs.\ 9.51), halving priming, and with 120 epochs it surpasses the 200-epoch baseline in all metrics. The relative gain is largest at small budgets ($-16\%$ \fdr{5} at 10 epochs), where the baseline has learned the least semantics, but it persists at 200 epochs. Alignment thus does not merely reach the same model sooner; it also improves the model that longer priming reaches.

\noindent\textbf{The speed-up survives a compute-matched comparison.} The teacher makes each B priming update 43\% more expensive, so faster progress per update need not mean faster training. It does here (\cref{fig:budget}c): \method-B with 80 epochs uses 2.5\% \emph{less} total compute than Priming-B with 120 epochs (102.3 vs.\ 105.0 EFLOPs) and is better in FID (1.404 vs.\ 1.415), \fdr{5} (11.24 vs.\ 11.82), and \fdr{6} (9.43 vs.\ 9.97).

\noindent\textbf{Fewer post-training updates.} If post-training builds on the interface formed during priming (\cref{sec:mechanism}), a better initialization should also shorten the expensive OT stage, and it does (\cref{fig:budget}d). From the 160-epoch initializations, \method reaches the baseline's final \fdr{6} (9.51) after 3.75k updates and surpasses it after 6.25k, $4\times$ fewer. At 40 epochs, \method after 2.5k updates equals the baseline after 22.5k (online \fdr{6} 13.71), a $9\times$ reduction. As an OT update costs ${\sim}4.9$ priming updates, this is the largest practical saving. The advantage is not a head start that longer post-training erases: at 50k updates the gap widens from 0.90 to 1.11 \fdr{6} (8.04 vs.\ 9.14), so post-training keeps exploiting, rather than replacing, what priming has learned.

\subsection{What to align with, where, and when}
\label{sec:ablation}

\begin{table}[t]
\centering
\footnotesize
\setlength{\tabcolsep}{3.5pt}
\begin{tabular}{@{}l l c c c@{}}
\toprule
& Setting & Align cos. & \fdr{6}$\downarrow$ & Recall$\uparrow$\\
\midrule
\multicolumn{5}{@{}l}{\textit{40-epoch priming, 2.5k post-training updates (EMA)}}\\
& no alignment & -- & 17.71 & .416\\
\multirow{3}{*}{\rotatebox{90}{\scriptsize teacher}} & DINOv2-L (cross-space) & 0.55 & \textbf{14.79} & .427\\
& ResNet-MAE (OT space) & 0.80 & 16.65 & .428\\
& ConvNeXt (OT space) & 0.97 & 17.49 & .412\\
\cmidrule(l){2-5}
\multirow{3}{*}{\rotatebox{90}{\scriptsize window}} & first 25\% of priming & -- & 15.89 & .422\\
& first 50\% of priming & -- & 15.52 & .426\\
& full priming & -- & \textbf{14.79} & .427\\
\cmidrule(l){2-5}
\multirow{4}{*}{\rotatebox{90}{\scriptsize depth}} & block 2 / 16 & -- & 15.78 & .413\\
& block 5 / 16 & -- & 14.79 & .427\\
& block 8 / 16 & -- & \textbf{14.54} & \textbf{.457}\\
& block 11 / 16 & -- & 14.93 & .449\\
\midrule
\multicolumn{5}{@{}l}{\textit{160/200-epoch priming, 25k post-training updates (EMA)}}\\
\multirow{4}{*}{\rotatebox{90}{\scriptsize teacher}} & no alignment & -- & 9.51 & .620\\
& DINOv2-L & -- & \textbf{8.61} & \textbf{.639}\\
& ResNet-MAE & -- & 9.65 & .622\\
& ConvNeXt & -- & 9.95 & .619\\
\cmidrule(l){2-5}
\multirow{2}{*}{\rotatebox{90}{\scriptsize depth}} & block 5, 160 / 200 ep. & -- & 8.61 / 8.19 & .639 / .631\\
& block 8, 160 / 200 ep. & -- & \textbf{8.28} / \textbf{8.15} & .641 / .635\\
\bottomrule
\end{tabular}
\caption{\textbf{Ablations of \method-B.} ``Align cos.'': training alignment cosine at the end of priming.}
\label{tab:ablation}
\vspace{-6pt}
\end{table}

\noindent\textbf{Teacher: it must add what the OT stage lacks.} \cref{tab:ablation} compares DINOv2 with the two feature spaces that the OT loss already uses. At 40 epochs, DINOv2 improves \fdr{6} by 16.5\%, ResNet-MAE by 6.0\%, and ConvNeXt by only 1.3\%; at full budget, MAE and ConvNeXt alignment even slightly hurt (9.65 and 9.95 vs.\ 9.51), whereas DINOv2 helps (8.61). The per-space breakdown (supplement) explains why. ConvNeXt alignment improves the ConvNeXt distance by 17\% but worsens CLIP by 8\%, \ie, it reinforces what post-training already optimizes. DINOv2 alignment instead improves CLIP and SigLIP, which neither training stage nor the teacher uses, by 18--20\%. Ease of alignment is \emph{inversely} related to benefit (final cosine 0.97, 0.80, and 0.55): a target that the projector matches almost perfectly exerts little pressure on the generator's features. A useful teacher must be semantic and lie outside the OT spaces; then alignment complements post-training instead of duplicating it.

\noindent\textbf{Window: do not stop early.} In diffusion, stopping REPA early helps~\cite{haste}. In priming, the benefit instead grows monotonically with the window (\fdr{6} 15.89, 15.52, and 14.79 for 25\%, 50\%, and full). Plausibly, priming trains at a single high noise level, where the semantic target stays informative throughout.

\noindent\textbf{Depth: mid-network.} Quality follows an inverted U over depth, with block 8 of 16 best at 40 epochs in both \fdr{6} and recall. Plausibly, early blocks are too shallow for semantics and late blocks serve pixel decoding. The ranking carries over to full budgets, where block 8 reaches 8.28 at 160 and 8.15 at 200 epochs, the best \fdr{6} with 25k post-training updates; our block-5 default thus understates the attainable gain.

\subsection{Joint single-stage training}
\label{sec:joint}

\begin{table}[t]
\centering
\footnotesize
\setlength{\tabcolsep}{2.6pt}
\begin{tabular}{@{}l r cc ccc@{}}
\toprule
& & \multicolumn{2}{c}{native} & \multicolumn{3}{c}{+25k OT updates}\\
\cmidrule(lr){3-4}\cmidrule(l){5-7}
Recipe & EFLOPs & FID & \fdr{5} & FID & \fdr{5} & \fdr{6}\\
\midrule
Priming-B (two-stage) & 124.2 & -- & -- & 1.373 & 11.27 & 9.51\\
\method-B (two-stage) & 157.3 & -- & -- & 1.303 & 10.20 & 8.61\\
Joint-B & 313.6 & \textbf{2.06} & 11.00 & 1.265 & 11.52 & 9.72\\
\rowcolor{rowgray}\methodJ-B & 346.8 & 2.47 & \textbf{9.75} & \textbf{1.224} & 10.60 & 8.90\\
\bottomrule
\end{tabular}
\caption{\textbf{Single-stage training} (B, 160 epochs, \ie, 200k joint updates). ``Native'': no post-training (EMA, one seed). EFLOPs include refinement and teacher.}
\label{tab:joint}
\vspace{-6pt}
\end{table}

\noindent\textbf{Why train both paths at once.} In the two-stage recipe, priming writes semantics into an interface and post-training teaches noise to use it (\cref{sec:mechanism}). Since post-training moves the weights by only 2--4\%, the generative path mostly needs to \emph{find} the interface, and nothing requires it to wait until priming ends: noise could learn to use the interface while it forms, yielding a \emph{native} one-step generator. Joint training (\cref{eq:joint}) tests this and whether the placement principle depends on the schedule. Since OT now shapes the generative path from the start, it might supply the semantics itself and make alignment redundant; \cref{prop:align} instead predicts that paired-path alignment still helps.

\noindent\textbf{Alignment remains essential when both paths train together.} In a $2\!\times\!2$ factorial at 40 epochs, each followed by a short 1.25k-update OT probe, alignment improves the sequential recipe from 18.11 to 14.61 \fdr{6} and the joint recipe from 15.11 to 12.44 (supplement). Joint training alone also helps (18.11$\to$15.11), and combining both is best. The effects are not redundant: the alignment gain shrinks only from 3.50 to 2.67, so early OT does not supply the teacher's semantics. The paired path, through alignment, determines \emph{what} the interface encodes; the generative path determines \emph{how} noise reaches it.

\noindent\textbf{Native one-step generation.} At full budget (\cref{tab:joint}), the native \methodJ reaches \fdr{5} 9.75 after 200k joint updates and 9.33 after 240k, better than every two-stage B model (best 9.66) without any post-training. Native \methodJ is ahead of native joint training in \fdr{5} at every checkpoint from 50k to 240k updates (13.96 vs.\ 16.71 at 50k), and it passes the two-stage Priming-B (11.27) after about 110k updates, whereas joint training without alignment needs about 190k. Native FID shows the opposite sign (2.47 vs.\ 2.06 at 200k), but the gap closes steadily, from 1.16 at 50k to 0.17 at 240k. We read this as slower convergence of low-level statistics, to which Inception is sensitive, while the semantics measured by \fdr{5} improve throughout.

\noindent\textbf{Refinement and cost.} Refining with 25k OT updates gives the best B-scale FID (1.224 vs.\ 1.265 without alignment) and lowers \fdr{6} by 8.5\% (9.72$\to$8.90), the same direction and similar size as in the two-stage recipe (9.51$\to$8.61). Refinement raises \fdr{5} over the native model (9.75$\to$10.60), so OT in ConvNeXt/MAE space trades some held-out semantics for fidelity. At $2.8\times$ the cost of Priming-B and $2.2\times$ that of \method-B, \methodJ is not the cheaper recipe; under a fixed budget, two-stage \method is the better choice. Its role is to show that \emph{align the pair, not the sample} holds even when the generative path is always active: the teacher's semantics reach noise through the shared weights, not through a target on the samples.

\section{Discussion and Conclusion}
\label{sec:conclusion}
\noindent\textbf{Limitations.} Our study covers class-conditional ImageNet at $256^2$, the Priming architecture, and OT post-training; text-to-image generation and other objectives remain to be tested. Most configurations are trained once, but re-runs (\cref{sec:main}) and a consistent sign across seven budgets suggest robust conclusions. Our generative-path targets all lack instance correspondence; objectives that create their own, \eg, trajectory-consistency views as in SF$^2$~\cite{sf2}, remain open. Finally, the teacher adds compute; cached features would reduce it.

\noindent\textbf{Conclusion.} Representation alignment helps a one-step generator on only one of its two paths. On the paired path, it writes image semantics into an interface through which post-training routes noise, halving the priming epochs and quartering the OT updates; on the generative path, every variant we tried trades coverage for typicality. We hope that \emph{align the pair, not the sample} will guide how representation learning and distribution matching are combined.

\section*{LLM Usage Statement}
LLM assistants were used to cross-check numbers, write plotting code, and edit the text. The authors designed and ran all experiments, verified all numbers, and take full responsibility for the content.

{\small
\bibliographystyle{ieeenat_fullname}
\bibliography{references}

@article{priming,
  author = {Park, Woobin and Kim, Jaeyeon and Du, Yilun and Chen, Sitan and Han, Bohyung},
  title = {{Are We Overthinking One-Step Generation?}},
  journal = {arXiv preprint arXiv:2610.20331},
  year = {2026},
  eprint = {2610.20331v2},
  archivePrefix = {arXiv},
  primaryClass = {cs.LG},
  url = {https://arxiv.org/abs/2610.20331v2}
}

@inproceedings{repa,
  author = {Yu, Sihyun and Kwak, Sangkyung and Jang, Huiwon and Jeong, Jongheon and Huang, Jonathan and Shin, Jinwoo and Xie, Saining},
  title = {{Representation Alignment for Generation: Training Diffusion Transformers Is Easier Than You Think}},
  booktitle = {International Conference on Learning Representations},
  year = {2025},
  eprint = {2410.06940v4},
  archivePrefix = {arXiv},
  url = {https://arxiv.org/abs/2410.06940v4}
}

@article{irepa,
  author = {Singh, Jaskirat and Leng, Xingjian and Wu, Zongze and Zheng, Liang and Zhang, Richard and Shechtman, Eli and Xie, Saining},
  title = {{What matters for Representation Alignment: Global Information or Spatial Structure?}},
  journal = {arXiv preprint arXiv:2512.10794},
  year = {2025},
  eprint = {2512.10794v1},
  archivePrefix = {arXiv},
  primaryClass = {cs.CV},
  url = {https://arxiv.org/abs/2512.10794v1}
}

@article{haste,
  author = {Wang, Ziqiao and Zhao, Wangbo and Zhou, Yuhao and Li, Zekai and Liang, Zhiyuan and Shi, Mingjia and Zhao, Xuanlei and Zhou, Pengfei and Zhang, Kaipeng and Wang, Zhangyang and Wang, Kai and You, Yang},
  title = {{REPA Works Until It Doesn't: Early-Stopped, Holistic Alignment Supercharges Diffusion Training}},
  journal = {arXiv preprint arXiv:2505.16792},
  year = {2025},
  eprint = {2505.16792v1},
  archivePrefix = {arXiv},
  primaryClass = {cs.CV},
  url = {https://arxiv.org/abs/2505.16792v1}
}

@article{semantictransport,
  author = {Van Assel, Hugues and De Brouwer, Edward and Saremi, Saeed and Scalia, Gabriele and Regev, Aviv},
  title = {{Generate in Reconstruction Space, Match in Semantic Space: Transport Geometry for One-Step Generation}},
  journal = {arXiv preprint arXiv:2606.00514},
  year = {2026},
  eprint = {2606.00514v2},
  archivePrefix = {arXiv},
  primaryClass = {cs.LG},
  url = {https://arxiv.org/abs/2606.00514v2}
}

@article{sf2,
  author = {Chen, Junyi and Zhang, Haiyu and Fu, Zhoujie and Wang, Tengfei and Guo, Chunchao},
  title = {{Beyond Distribution Matching: Self-Supervised Representation Forcing for Few-Step Video Generation}},
  journal = {arXiv preprint arXiv:2610.35594},
  year = {2026},
  eprint = {2610.35594v1},
  archivePrefix = {arXiv},
  primaryClass = {cs.CV},
  url = {https://arxiv.org/abs/2610.35594v1}
}

@article{easd,
  author = {Peng, Fei and Wu, Junqiang and Tan, Haoxian and Zhou, Jun and Hu, Jie and Wei, Xiaoming and Guo, Jie and Li, Xiu},
  title = {{Efficient Hybrid Distillation: Synergizing Score and Adversarial Objectives for One-Step Diffusion}},
  journal = {arXiv preprint arXiv:2611.03017},
  year = {2026},
  eprint = {2611.03017v1},
  archivePrefix = {arXiv},
  primaryClass = {cs.CV},
  url = {https://arxiv.org/abs/2611.03017v1}
}

@article{drifting,
  author = {Deng, Mingyang and Li, He and Li, Tianhong and Du, Yilun and He, Kaiming},
  title = {{Generative Modeling via Drifting}},
  journal = {arXiv preprint arXiv:2602.04770},
  year = {2026},
  eprint = {2602.04770v2},
  archivePrefix = {arXiv},
  primaryClass = {cs.LG},
  url = {https://arxiv.org/abs/2602.04770v2}
}

@inproceedings{dmd,
  author = {Yin, Tianwei and Gharbi, Micha{\"e}l and Zhang, Richard and Shechtman, Eli and Durand, Fredo and Freeman, William T. and Park, Taesung},
  title = {{One-step Diffusion with Distribution Matching Distillation}},
  booktitle = {Proceedings of the IEEE/CVF Conference on Computer Vision and Pattern Recognition},
  year = {2024},
  eprint = {2311.18828v4},
  archivePrefix = {arXiv},
  url = {https://arxiv.org/abs/2311.18828v4}
}

@article{dmd2,
  author = {Yin, Tianwei and Gharbi, Micha{\"e}l and Park, Taesung and Zhang, Richard and Shechtman, Eli and Durand, Fredo and Freeman, William T.},
  title = {{Improved Distribution Matching Distillation for Fast Image Synthesis}},
  journal = {arXiv preprint arXiv:2405.14867},
  year = {2024},
  eprint = {2405.14867v2},
  archivePrefix = {arXiv},
  primaryClass = {cs.CV},
  url = {https://arxiv.org/abs/2405.14867v2}
}

@article{meanflow,
  author = {Geng, Zhengyang and Deng, Mingyang and Bai, Xingjian and Kolter, J. Zico and He, Kaiming},
  title = {{Mean Flows for One-step Generative Modeling}},
  journal = {arXiv preprint arXiv:2505.13447},
  year = {2025},
  eprint = {2505.13447v1},
  archivePrefix = {arXiv},
  primaryClass = {cs.LG},
  url = {https://arxiv.org/abs/2505.13447v1}
}

@article{dinov2,
  author = {Oquab, Maxime and Darcet, Timoth{\'e}e and Moutakanni, Th{\'e}o and Vo, Huy and Szafraniec, Marc and Khalidov, Vasil and Fernandez, Pierre and Haziza, Daniel and Massa, Francisco and El-Nouby, Alaaeldin and Assran, Mahmoud and Ballas, Nicolas and Galuba, Wojciech and Howes, Russell and Huang, Po-Yao and Li, Shang-Wen and Misra, Ishan and Rabbat, Michael and Sharma, Vasu and Synnaeve, Gabriel and Xu, Hu and Jegou, Herv{\'e} and Mairal, Julien and Labatut, Patrick and Joulin, Armand and Bojanowski, Piotr},
  title = {{DINOv2: Learning Robust Visual Features without Supervision}},
  journal = {arXiv preprint arXiv:2304.07193},
  year = {2023},
  eprint = {2304.07193v2},
  archivePrefix = {arXiv},
  primaryClass = {cs.CV},
  url = {https://arxiv.org/abs/2304.07193v2}
}

@inproceedings{convnext,
  author = {Liu, Zhuang and Mao, Hanzi and Wu, Chao-Yuan and Feichtenhofer, Christoph and Darrell, Trevor and Xie, Saining},
  title = {{A ConvNet for the 2020s}},
  booktitle = {Proceedings of the IEEE/CVF Conference on Computer Vision and Pattern Recognition},
  year = {2022},
  eprint = {2201.03545v2},
  archivePrefix = {arXiv},
  url = {https://arxiv.org/abs/2201.03545v2}
}

@inproceedings{mae,
  author = {He, Kaiming and Chen, Xinlei and Xie, Saining and Li, Yanghao and Doll{\'a}r, Piotr and Girshick, Ross},
  title = {{Masked Autoencoders Are Scalable Vision Learners}},
  booktitle = {Proceedings of the IEEE/CVF Conference on Computer Vision and Pattern Recognition},
  year = {2022},
  pages = {16000--16009},
  eprint = {2111.06377v3},
  archivePrefix = {arXiv},
  url = {https://openaccess.thecvf.com/content/CVPR2022/html/He_Masked_Autoencoders_Are_Scalable_Vision_Learners_CVPR_2022_paper.html}
}

@inproceedings{siglip,
  author = {Zhai, Xiaohua and Mustafa, Basil and Kolesnikov, Alexander and Beyer, Lucas},
  title = {{Sigmoid Loss for Language Image Pre-Training}},
  booktitle = {Proceedings of the IEEE/CVF International Conference on Computer Vision},
  year = {2023},
  eprint = {2303.15343v4},
  archivePrefix = {arXiv},
  url = {https://arxiv.org/abs/2303.15343v4}
}

@inproceedings{clip,
  author = {Radford, Alec and Kim, Jong Wook and Hallacy, Chris and Ramesh, Aditya and Goh, Gabriel and Agarwal, Sandhini and Sastry, Girish and Askell, Amanda and Mishkin, Pamela and Clark, Jack and Krueger, Gretchen and Sutskever, Ilya},
  title = {{Learning Transferable Visual Models From Natural Language Supervision}},
  booktitle = {Proceedings of the 38th International Conference on Machine Learning},
  series = {Proceedings of Machine Learning Research},
  volume = {139},
  pages = {8748--8763},
  publisher = {PMLR},
  year = {2021},
  eprint = {2103.00020v1},
  archivePrefix = {arXiv},
  url = {https://proceedings.mlr.press/v139/radford21a.html}
}

@inproceedings{fid,
  author = {Heusel, Martin and Ramsauer, Hubert and Unterthiner, Thomas and Nessler, Bernhard and Hochreiter, Sepp},
  title = {{GANs Trained by a Two Time-Scale Update Rule Converge to a Local Nash Equilibrium}},
  booktitle = {Advances in Neural Information Processing Systems},
  volume = {30},
  year = {2017},
  eprint = {1706.08500v6},
  archivePrefix = {arXiv},
  url = {https://arxiv.org/abs/1706.08500v6}
}

@inproceedings{precisionrecall,
  author = {Kynk{\"a}{\"a}nniemi, Tuomas and Karras, Tero and Laine, Samuli and Lehtinen, Jaakko and Aila, Timo},
  title = {{Improved Precision and Recall Metric for Assessing Generative Models}},
  booktitle = {Advances in Neural Information Processing Systems},
  year = {2019},
  eprint = {1904.06991v3},
  archivePrefix = {arXiv},
  url = {https://arxiv.org/abs/1904.06991v3}
}

@inproceedings{sinkhorn,
  author = {Cuturi, Marco},
  title = {{Sinkhorn Distances: Lightspeed Computation of Optimal Transportation Distances}},
  booktitle = {Advances in Neural Information Processing Systems},
  volume = {26},
  pages = {2292--2300},
  year = {2013},
  eprint = {1306.0895v1},
  archivePrefix = {arXiv},
  url = {https://arxiv.org/abs/1306.0895v1}
}

@inproceedings{sinkhorndivergences,
  author = {Feydy, Jean and S{\'e}journ{\'e}, Thibault and Vialard, Fran{\c c}ois-Xavier and Amari, Shun-ichi and Trouv{\'e}, Alain and Peyr{\'e}, Gabriel},
  title = {{Interpolating between Optimal Transport and MMD using Sinkhorn Divergences}},
  booktitle = {Proceedings of the Twenty-Second International Conference on Artificial Intelligence and Statistics},
  series = {Proceedings of Machine Learning Research},
  volume = {89},
  pages = {2681--2690},
  publisher = {PMLR},
  year = {2019},
  eprint = {1810.08278v1},
  archivePrefix = {arXiv},
  url = {https://proceedings.mlr.press/v89/feydy19a.html}
}

@article{fdloss,
  author = {Yang, Jiawei and Geng, Zhengyang and Ju, Xuan and Tian, Yonglong and Wang, Yue},
  title = {{Representation Fr{\'e}chet Loss for Visual Generation}},
  journal = {arXiv preprint arXiv:2604.28190},
  year = {2026},
  eprint = {2604.28190v1},
  archivePrefix = {arXiv},
  primaryClass = {cs.CV},
  url = {https://arxiv.org/abs/2604.28190v1}
}

@inproceedings{imagenet,
  author = {Deng, Jia and Dong, Wei and Socher, Richard and Li, Li-Jia and Li, Kai and Fei-Fei, Li},
  title = {{ImageNet: A large-scale hierarchical image database}},
  booktitle = {2009 IEEE Conference on Computer Vision and Pattern Recognition},
  pages = {248--255},
  year = {2009},
  doi = {10.1109/CVPR.2009.5206848},
  url = {https://doi.org/10.1109/CVPR.2009.5206848}
}

@inproceedings{lpips,
  author = {Zhang, Richard and Isola, Phillip and Efros, Alexei A. and Shechtman, Eli and Wang, Oliver},
  title = {{The Unreasonable Effectiveness of Deep Features as a Perceptual Metric}},
  booktitle = {Proceedings of the IEEE/CVF Conference on Computer Vision and Pattern Recognition},
  year = {2018},
  eprint = {1801.03924v2},
  archivePrefix = {arXiv},
  url = {https://arxiv.org/abs/1801.03924v2}
}

@article{wflow,
  author = {Han, Jiaqi and Li, Puheng and Guo, Qiushan and Xu, Renyuan and Ermon, Stefano and Cand{\`e}s, Emmanuel J.},
  title = {{One-Step Generative Modeling via Wasserstein Gradient Flows}},
  journal = {arXiv preprint arXiv:2605.11755},
  year = {2026}
}

@inproceedings{pmf,
  author = {Lu, Yiyang and Lu, Susie and Sun, Qiao and Zhao, Hanhong and Jiang, Zhicheng and Wang, Xianbang and Li, Tianhong and Geng, Zhengyang and He, Kaiming},
  title = {{One-step Latent-free Image Generation with Pixel Mean Flows}},
  booktitle = {Proceedings of the 43rd International Conference on Machine Learning},
  series = {Proceedings of Machine Learning Research},
  volume = {306},
  pages = {81265--81277},
  year = {2026}
}

@inproceedings{imf,
  author = {Geng, Zhengyang and Lu, Yiyang and Wu, Zongze and Shechtman, Eli and Kolter, J. Zico and He, Kaiming},
  title = {{Improved Mean Flows: On the Challenges of Fastforward Generative Models}},
  booktitle = {Proceedings of the IEEE/CVF Conference on Computer Vision and Pattern Recognition},
  year = {2026}
}

@article{dispersive,
  author = {Wang, Runqian and He, Kaiming},
  title = {{Diffuse and Disperse: Image Generation with Representation Regularization}},
  journal = {arXiv preprint arXiv:2506.09027},
  year = {2025}
}

@inproceedings{jit,
  author = {Li, Tianhong and He, Kaiming},
  title = {{Back to Basics: Let Denoising Generative Models Denoise}},
  booktitle = {Proceedings of the IEEE/CVF Conference on Computer Vision and Pattern Recognition},
  year = {2026}
}

@inproceedings{consistency,
  author = {Song, Yang and Dhariwal, Prafulla and Chen, Mark and Sutskever, Ilya},
  title = {{Consistency Models}},
  booktitle = {Proceedings of the 40th International Conference on Machine Learning},
  year = {2023}
}

@inproceedings{shortcut,
  author = {Frans, Kevin and Hafner, Danijar and Levine, Sergey and Abbeel, Pieter},
  title = {{One Step Diffusion via Shortcut Models}},
  booktitle = {International Conference on Learning Representations},
  year = {2025}
}

@inproceedings{imm,
  author = {Zhou, Linqi and Ermon, Stefano and Song, Jiaming},
  title = {{Inductive Moment Matching}},
  booktitle = {Proceedings of the 42nd International Conference on Machine Learning},
  year = {2025}
}

@inproceedings{gmmn,
  author = {Li, Yujia and Swersky, Kevin and Zemel, Richard},
  title = {{Generative Moment Matching Networks}},
  booktitle = {Proceedings of the 32nd International Conference on Machine Learning},
  year = {2015}
}

@inproceedings{genevay2018,
  author = {Genevay, Aude and Peyr{\'e}, Gabriel and Cuturi, Marco},
  title = {{Learning Generative Models with Sinkhorn Divergences}},
  booktitle = {Proceedings of the Twenty-First International Conference on Artificial Intelligence and Statistics},
  year = {2018}
}

@inproceedings{repae,
  author = {Leng, Xingjian and Singh, Jaskirat and Hou, Yunzhong and Xing, Zhenchang and Xie, Saining and Zheng, Liang},
  title = {{REPA-E: Unlocking VAE for End-to-End Tuning with Latent Diffusion Transformers}},
  booktitle = {Proceedings of the IEEE/CVF International Conference on Computer Vision},
  year = {2025}
}

@article{rae,
  author = {Zheng, Boyang and Ma, Nanye and Tong, Shengbang and Xie, Saining},
  title = {{Diffusion Transformers with Representation Autoencoders}},
  journal = {arXiv preprint arXiv:2510.11690},
  year = {2025}
}

@inproceedings{vavae,
  author = {Yao, Jingfeng and Yang, Bin and Wang, Xinggang},
  title = {{Reconstruction vs. Generation: Taming Optimization Dilemma in Latent Diffusion Models}},
  booktitle = {Proceedings of the IEEE/CVF Conference on Computer Vision and Pattern Recognition},
  year = {2025}
}

@inproceedings{dit,
  author = {Peebles, William and Xie, Saining},
  title = {{Scalable Diffusion Models with Transformers}},
  booktitle = {Proceedings of the IEEE/CVF International Conference on Computer Vision},
  year = {2023}
}

@inproceedings{ning2024exposure,
  author = {Ning, Mang and Li, Mingxiao and Su, Jianlin and Salah, Albert Ali and Ertugrul, Itir Onal},
  title = {{Elucidating the Exposure Bias in Diffusion Models}},
  booktitle = {International Conference on Learning Representations},
  year = {2024}
}

@inproceedings{li2024shifted,
  author = {Li, Mingxiao and Qu, Tingyu and Yao, Ruicong and Sun, Wei and Moens, Marie-Francine},
  title = {{Alleviating Exposure Bias in Diffusion Models through Sampling with Shifted Time Steps}},
  booktitle = {International Conference on Learning Representations},
  year = {2024}
}

@inproceedings{dctdiff,
  author = {Ning, Mang and Li, Mingxiao and Su, Jianlin and Jia, Haozhe and Liu, Lanmiao and Bene{\v{s}}, Martin and Chen, Wenshuo and Salah, Albert Ali and Ertugrul, Itir Onal},
  title = {{DCTdiff: Intriguing Properties of Image Generative Modeling in the DCT Space}},
  booktitle = {Proceedings of the 42nd International Conference on Machine Learning},
  year = {2025}
}

@article{spectrummatching,
  author = {Ning, Mang and Li, Mingxiao and Zhang, Le and Liu, Lanmiao and Blaschko, Matthew B. and Salah, Albert Ali and Ertugrul, Itir Onal},
  title = {{Spectrum Matching: A Unified Perspective for Superior Diffusability in Latent Diffusion}},
  journal = {arXiv preprint arXiv:2603.14645},
  year = {2026}
}

@inproceedings{zigzagstory,
  author = {Li, Mingxiao and Ning, Mang and Moens, Marie-Francine},
  title = {{Consistent Story Generation: Unlocking the Potential of Zigzag Sampling}},
  booktitle = {Advances in Neural Information Processing Systems},
  year = {2025}
}

@inproceedings{animatemotion,
  author = {Li, Mingxiao and Wan, Bo and Moens, Marie-Francine and Tuytelaars, Tinne},
  title = {{Animate Your Motion: Turning Still Images into Dynamic Videos}},
  booktitle = {European Conference on Computer Vision},
  year = {2024}
}

@article{li2025personalized,
  author = {Li, Mingxiao and Qu, Tingyu and Tuytelaars, Tinne and Moens, Marie-Francine},
  title = {{Towards More Accurate Personalized Image Generation: Addressing Overfitting and Evaluation Bias}},
  journal = {arXiv preprint arXiv:2503.06632},
  year = {2025}
}
}
\clearpage
\appendix
% ---------------------------------------------------------------------------
% Supplementary material body for "Align the Pair, Not the Sample".
% This file is a pure fragment with NO class declaration: main.tex inputs it right after
% \appendix, so the whole paper + supplement is a single arXiv document.
% (The old supp.tex used the subfiles package and could not compile on its own,
%  because it re-loaded preamble.tex and hit duplicate macro-definition errors.)
% ---------------------------------------------------------------------------
\renewcommand{\thesection}{\Alph{section}}
\renewcommand{\thetable}{S\arabic{table}}
\renewcommand{\thefigure}{S\arabic{figure}}
\renewcommand{\theequation}{S\arabic{equation}}
% restart the counters so the supplement numbers its own tables/figures/equations
% S1, S2, ... (the counters otherwise continue from the main paper)
\setcounter{table}{0}
\setcounter{figure}{0}
\setcounter{equation}{0}

\twocolumn[{\suppTitle}]

\noindent This supplement contains the proof of Proposition~1 (\cref{sec:s-proof}), implementation and evaluation details (\cref{sec:s-impl,sec:s-eval}), compute accounting (\cref{sec:s-compute}), complete result tables (\cref{sec:s-results}), additional analyses (\cref{sec:s-analysis}), and uncurated samples (\cref{sec:s-samples}). All numbers are read directly from the stored evaluation outputs of each run; scripts that regenerate every table and figure from these outputs are part of the code release.

\section{Proof of Proposition 1}
\label{sec:s-proof}
Let $X$ be the generator input with class $c$, $T\in\mathbb S^{d-1}$ the target token, and $u:\mathcal X\to\mathbb S^{d-1}$ any measurable prediction. Write $\mu(X)=\E[T\mid X]$ and $m_c=\E[T\mid c]$.

\noindent\textit{(i)} By the tower property,
$\mathcal R(u)=1-\E\big[u(X)^\top\E[T\mid X]\big]=1-\E[u(X)^\top\mu(X)].$
Since $\|u(X)\|=1$, Cauchy--Schwarz gives $u(X)^\top\mu(X)\le\|\mu(X)\|$ pointwise, with equality iff $u(X)=\mu(X)/\|\mu(X)\|$ whenever $\mu(X)\neq0$ (any unit vector is optimal where $\mu(X)=0$). Taking expectations yields $\mathcal R(u)\ge1-\E\|\mu(X)\|$, attained by $u^\star$.

\noindent\textit{(ii)} If $T\perp X\mid c$, then $\mu(X)=\E[T\mid X]=\E[T\mid c]=m_c$, which depends on $X$ only through $c$. By (i), the optimal prediction is $u^\star(X)=m_c/\|m_c\|$ for all $X$ of class $c$ (if $m_c\ne0$), with risk $1-\E_c\|m_c\|$.

\noindent\textit{(iii)} Since $c$ is a function of $X$ (the class label is an input), $m_c=\E[\mu(X)\mid c]$. The norm is convex, so by conditional Jensen $\E[\|\mu(X)\|\mid c]\ge\|\E[\mu(X)\mid c]\|=\|m_c\|$. Equality holds iff, within each class, the vectors $\mu(X)$ are almost surely non-negative multiples of a common direction, \ie, iff $u^\star$ is constant within each class. Averaging over $c$ gives $\mathcal A\ge0$ with the stated equality condition. \hfill$\square$

\noindent\textbf{Remarks.} (1) The proposition concerns the population optimum of the alignment term alone. In training, the alignment loss is combined with reconstruction or OT losses and acts through a projector $P_\psi$. Its \emph{gradient} with respect to the generator features nevertheless points, for unpaired targets, toward a feature configuration that is class-constant after projection; this is the attractor discussed in the main paper. (2) For max-over-tokens cosine (used in 10 of our generative-path configurations), $\mathcal R$ is replaced by $\E[1-\max_q u_p^\top T_q]$. With random or pooled targets, the target set is again independent of $X$ given $c$, so the minimizer is again class-constant. We measured how close a constant prediction comes, using DINOv2 tokens of 1024 real ImageNet images. A single class-mean direction reaches a max-cosine of 0.60, which is optimistic because it is fitted in-sample. A trained linear probe with per-image token selection reaches 0.63, and a single global direction 0.37. Most of the attainable max-cosine is thus available to a class-constant predictor. (3) For the paired path, $\mathcal A>0$ is an empirical statement. On the released Priming-B (160 epochs, no alignment), a ridge probe from block-8 features of $x_s$ ($s{=}0.8$) recovers the clean image's DINOv2 tokens with a per-token cosine of 0.36. The cosine drops to 0.18 when the image pairing is shuffled. Relative to this shuffled floor, the generator features carry $+0.185$ of recoverable target structure, compared with $+0.041$ for a probe on the raw pixels of $x_s$ (80/20 image-level split, bootstrap std.\ 0.004).

\section{Implementation details}
\label{sec:s-impl}
\begin{table}[h]
\centering\footnotesize
\setlength{\tabcolsep}{3pt}
\begin{tabular}{@{}p{0.2\linewidth}p{0.77\linewidth}@{}}
\toprule
\multicolumn{2}{@{}l}{\textit{Generator (from~\cite{priming})}}\\
B / L & 16 / 32 blocks, width 768 / 1024; 115.1M / 405.1M params\\
Tokens & patch 16 ($16\times16$ grid) + 8 class tokens, 2D RoPE\\
\midrule
\multicolumn{2}{@{}l}{\textit{Priming (Stage 1)}}\\
Corruption & $x_s=0.2\,x_0+0.8\,\epsilon$ (fixed $s=0.8$)\\
Loss & adaptive pixel MSE + 0.4 LPIPS + 0.1 ConvNeXt\\
Optimizer & Muon (mom.\ 0.95), lr $10^{-3}$ const., no WD\\
Batch / EMA & 1024 / 0.9999; bf16\\
Epochs & 160 (200,160 updates); budgets 10--200\\
Data aug. & center crop $256^2$, random horizontal flip\\
\midrule
\multicolumn{2}{@{}l}{\textit{\method alignment}}\\
Teacher & DINOv2-L/14 (\texttt{lvd142m}), frozen, FP32\\
Teacher input & $x_0$ resized to $224^2$ $\to$ $16\times16$ patch tokens\\
Generator feature & block 5 (B, default), 8 (B, best), 10 (L)\\
Projector & Linear($D$,1024)--GELU--Linear(1024,1024)\\
Projector opt. & AdamW, lr $10^{-4}$\\
Loss & tokenwise $1-\cos$, $\lambda=0.5$, whole priming\\
\midrule
\multicolumn{2}{@{}l}{\textit{Post-training (Stage 2, from~\cite{priming})}}\\
Updates & 25k (50k for the extended run)\\
Optimizer & AdamW, lr $10^{-5}$ (125-step warm-up), clip 1.0\\
EMA & 0.999\\
Batch & 16 classes $\times$ (64 + 64 sg) samples, 128 real/class\\
OT & $0.75\,\mathrm{OT}_{0.001}+0.25\,\mathrm{OT}_{0.01}$, log-Sinkhorn\\
Features & ConvNeXt + ResNet-MAE stack, coefficient 5\\
\midrule
\multicolumn{2}{@{}l}{\textit{\methodJ}}\\
OT frequency & every $k=2$ outer updates\\
OT weight & linear ramp $0\to1$ over the first 25k updates\\
Schedule & 200,160 joint updates (240,192 for the extended run);\\
& optional refinement: Stage 2 for 25k updates\\
\bottomrule
\end{tabular}
\caption{\textbf{Hyper-parameters.} Everything not listed under \method is identical for Priming and \method.}
\label{tab:s-hparams}
\end{table}

\noindent\textbf{Generative-path alignment.} All generative-path runs start from the released Priming-B 160-epoch initialization and use the standard post-training recipe for 2.5k updates. The alignment loss acts on block 8 through a projector of the same form as above. It is activated after 250 updates and ramped linearly over 100 updates, with $\lambda=1$ unless stated otherwise. Targets are (i)~DINOv2 tokens of a random real image of the same class from the OT reference set; (ii)~a pool of tokens from all same-class reference images, with max-over-tokens matching; (iii)~DINOv2 tokens of the reference image with the largest OT-plan mass for each sample; (iv)~DINOv2 tokens of the sample itself (stop-gradient); and (v)~tokens of an EMA teacher's sample for the same noise. For the ConvNeXt variants, the targets are the stage-2 tokens of the frozen OT ConvNeXt. Max-cos matches each predicted token to its most similar target token, whereas tokenwise matches tokens at the same position. All runs of this 2.5k-update campaign, including its no-alignment baseline and the paired-path counterpart, use an earlier export of the ImageNet training set with a different packing than the official one used in the full-budget experiments. We therefore compare them only with each other.

\noindent\textbf{Joint training.} OT steps use the post-training machinery inside the priming loop: one forward/backward pass on 16 classes $\times$ 64 samples (plus a stop-gradient negative batch) every second update. OT gradients are all-reduced separately and added to the reconstruction gradients before a single optimizer step. Joint runs were trained in segments (40 epochs, then continued to 160 and 192 epochs) with identical hyper-parameters. The logged OT weight is 1.0 after the initial ramp in every segment.

\section{Evaluation protocol}
\label{sec:s-eval}
\noindent\textbf{FID and \fdr{5}.} We use the official evaluation script of~\cite{priming} with 50k samples, one network evaluation, no guidance, and EMA weights. FID uses the JiT ImageNet-256 reference statistics~\cite{jit}. \fdr{5} averages Fr\'echet distances in Inception, DINOv2, ViT-MAE, SigLIP, and CLIP (CLS) features, each normalized by a validation reference value~\cite{fdloss,priming}. Reported values average three sampling seeds where stated. For \method-B at 160 epochs, the same checkpoint was evaluated twice with three seeds each, and we average all six evaluations.

\noindent\textbf{\fdr{6}, precision, recall.} The training-time evaluator computes six FDs (Inception/ADM, CLIP, ConvNeXt, DINOv2, MAE, SigLIP) on 50k samples and reports
\begin{equation}
\fdr{6}=\tfrac16\big(\tfrac{\mathrm{FD}_{\rm Inc}}{1.68}+\tfrac{\mathrm{FD}_{\rm CLIP}}{5.6}+\tfrac{\mathrm{FD}_{\rm CNX}}{56.87}+\tfrac{\mathrm{FD}_{\rm DINO}}{14.19}+\tfrac{\mathrm{FD}_{\rm MAE}}{0.04}+\tfrac{\mathrm{FD}_{\rm SigLIP}}{0.6}\big),
\end{equation}
together with precision and recall ($k=3$, Inception features, ADM ImageNet-256 reference batch). This evaluator runs every 1.25k post-training updates for both online and EMA weights. Its Inception FD is not the official FID.

\section{Compute accounting}
\label{sec:s-compute}
We follow the EFLOP accounting of~\cite{priming}. From the reported 85.75/124.18 EFLOPs for Priming-B at 80/160 epochs, one priming epoch costs 0.480 EFLOPs and post-training (25k updates) costs 47.3 EFLOPs. For L, these are 1.096 and 63.6 EFLOPs. A DINOv2-L/14 forward pass at $224^2$ (24 layers, width 1024, 257 tokens) costs $2\cdot24\,(12Td^2+2T^2d)=162$ GFLOPs per image, \ie, 0.207 EFLOPs per epoch. This adds 43\% (B) and 19\% (L) to a priming epoch. One OT update costs $47.3/25{,}000=1.89$ PFLOPs, \ie, about $4.9\times$ a B priming update, so a joint run with 100k OT updates adds 189 EFLOPs. Our implementation evaluates the teacher in FP32 without compilation. Measured on 8$\times$H200, the wall-clock time per B priming update is 1.23\,s for \method versus 0.66--0.72\,s for the compiled baseline. Teacher features depend only on the image and its flip, so they can be precomputed once.

\section{Complete results}
\label{sec:s-results}
This section lists the numbers behind the main paper's figures and summarizes what each table shows.

\noindent\textbf{Priming budget} (\cref{tab:s-budget}). \method is better than Priming at all seven budgets in every metric: FID, \fdr{5}, \fdr{6}, and recall. The relative \fdr{5} gain is largest at 10 epochs ($-16\%$) and remains $-11\%$ at 200 epochs. \method at 80 epochs matches Priming at 160 epochs (\fdr{5} 11.24 vs.\ 11.27), and \method at 120 epochs beats Priming at 200 epochs in all four metrics. Recall improves at every budget, so the gain does not come from concentrating samples. The teacher's cost moves each \method point to the right, but \method at 80 epochs still beats Priming at 120 epochs with 2.5\% less total compute.

\noindent\textbf{Generative-path configurations} (\cref{tab:s-generated}). Every row lowers recall well below the baseline (0.149--0.252 vs.\ 0.452, EMA). Precision stays at or above the baseline in 16 of 17 rows, and IS often rises. Samples become more typical, not individually worse, which is the prototype attraction predicted by Proposition~1. The pattern does not depend on the target construction, the target space, the loss form, the loss weight, or the alignment window. Alignment to the sample itself or to an EMA teacher behaves like alignment to a random image. Continuing the weakest run with 2.5k pure-OT updates restores only a small part of the lost recall.

\noindent\textbf{Per-space distances} (\cref{tab:s-components}). Our main claims should not depend on the teacher's own feature space. Without the DINOv2 column, the average still drops by 8.6\% (B-160), 10.8\% (B-200), and 12.7\% (L). At L, \method improves every space. At B, it improves all spaces except CLIP at 160 epochs, which is unchanged (22.14 vs.\ 22.15), and at 200 epochs CLIP improves as well. The largest relative gains appear in SigLIP (16--19\%), a space that neither training stage nor the teacher uses. After joint training and refinement, \methodJ improves every space except CLIP.

\noindent\textbf{Long post-training from 40 epochs} (\cref{tab:s-init40}). The \method initialization leads at every post-training length, and its lead in \fdr{6} narrows but persists (3.50 at 1.25k updates, 2.02 at 25k). \method after 2.5k updates (13.71) equals Priming after 22.5k updates, a $9\times$ saving in the most expensive stage. Recall is also higher at every step. Post-training therefore builds on what alignment wrote during priming instead of catching up from a weaker start.

\noindent\textbf{Joint training} (\cref{tab:s-native,tab:s-jointdepth}). Native \methodJ beats native joint training in \fdr{5} at every checkpoint, and its native FID gap shrinks steadily from 1.16 at 50k updates to 0.17 at 240k. Alignment depth trades off the two metrics: \fdr{}-type metrics prefer block 5, while native FID improves with depth. With the generative path trained from the start, a deeper interface plausibly leaves fewer blocks for the noise path to adapt before decoding pixels. Two-stage \method instead prefers block 8 in both \fdr{6} and recall (main paper, Tab.~2).
\begin{table*}[t]
\centering\footnotesize
\begin{tabular}{@{}r rcccc rcccc@{}}\toprule
& \multicolumn{5}{c}{Priming-B} & \multicolumn{5}{c}{\method-B}\\\cmidrule(lr){2-6}\cmidrule(l){7-11}
Ep. & EFLOPs & FID & \fdr{5} & \fdr{6} & Rec. & EFLOPs & FID & \fdr{5} & \fdr{6} & Rec.\\\midrule
10 & 52.1 & 2.551 & 21.63 & 18.27 & 0.556 & 54.2 & \textbf{2.157} & \textbf{18.23} & \textbf{15.39} & \textbf{0.576}\\
20 & 56.9 & 1.993 & 17.79 & 14.99 & 0.581 & 61.1 & \textbf{1.709} & \textbf{15.38} & \textbf{12.96} & \textbf{0.596}\\
40 & 66.5 & 1.703 & 15.22 & 12.80 & 0.604 & 74.8 & \textbf{1.507} & \textbf{12.96} & \textbf{10.93} & \textbf{0.614}\\
80 & 85.8 & 1.495 & 12.90 & 10.86 & 0.614 & 102.3 & \textbf{1.404} & \textbf{11.24} & \textbf{9.43} & \textbf{0.623}\\
120 & 105.0 & 1.415 & 11.82 & 9.97 & 0.619 & 129.8 & \textbf{1.349} & \textbf{10.76} & \textbf{9.02} & \textbf{0.627}\\
160 & 124.2 & 1.373 & 11.27 & 9.51 & 0.620 & 157.3 & \textbf{1.303} & \textbf{10.20} & \textbf{8.61} & \textbf{0.639}\\
200 & 143.4 & 1.353 & 10.91 & 9.21 & 0.625 & 184.8 & \textbf{1.283} & \textbf{9.66} & \textbf{8.19} & \textbf{0.631}\\
\bottomrule\end{tabular}
\caption{\textbf{Priming budget sweep (B).} Every model is post-trained for 25k updates. FID/\fdr{5} are single-seed except at 160 and 200 epochs, which use the same seed averages as Tab.~1 of the main paper (three seeds; six evaluations for \method at 160 epochs). \fdr{6} and recall use EMA weights. The 80-epoch Priming model is the released checkpoint; other Priming budgets come from our re-training of the official recipe (10--160 epochs: one run with milestones; 200 epochs: a separate run).}
\label{tab:s-budget}
\end{table*}

\begin{table*}[t]
\centering\footnotesize
\setlength{\tabcolsep}{4pt}
\begin{tabular}{@{}lllcc cccc cc@{}}\toprule
& & & & & \multicolumn{4}{c}{EMA} & \multicolumn{2}{c}{online}\\\cmidrule(lr){6-9}\cmidrule(l){10-11}
Target & Target space & Loss & $\lambda$ & Window & \fdr{6} & Rec. & Prec. & IS & \fdr{6} & Rec.\\\midrule
\multicolumn{5}{@{}l}{no alignment (baseline)} & 14.71 & 0.452 & 0.787 & 270 & 13.80 & 0.494\\\midrule
OT-matched real & DINOv2 & max-cos & 1 & 250--2.5k & 16.86 & 0.200 & 0.824 & 290 & 17.22 & 0.246\\
random same-class & DINOv2 & max-cos & 1 & 250--2.5k & 16.71 & 0.203 & 0.826 & 291 & 17.18 & 0.255\\
same-class token pool & DINOv2 & max-cos & 1 & 250--2.5k & 17.14 & 0.189 & 0.839 & 297 & 18.30 & 0.249\\
own sample (sg) & DINOv2 & max-cos & 1 & 250--2.5k & 16.90 & 0.201 & 0.820 & 286 & 17.78 & 0.257\\
EMA-teacher sample & DINOv2 & max-cos & 1 & 250--2.5k & 17.93 & 0.182 & 0.830 & 290 & 19.42 & 0.227\\
OT-matched real & ConvNeXt (OT) & max-cos & 1 & 250--2.5k & 16.28 & 0.195 & 0.835 & 297 & 16.02 & 0.256\\
random same-class & ConvNeXt (OT) & max-cos & 1 & 250--2.5k & 16.43 & 0.193 & 0.836 & 299 & 16.05 & 0.268\\
same-class token pool & ConvNeXt (OT) & max-cos & 1 & 250--2.5k & 16.87 & 0.181 & 0.840 & 301 & 16.55 & 0.256\\
own sample (sg) & ConvNeXt (OT) & max-cos & 1 & 250--2.5k & 16.46 & 0.192 & 0.830 & 299 & 16.11 & 0.261\\
EMA-teacher sample & ConvNeXt (OT) & max-cos & 1 & 250--2.5k & 16.56 & 0.186 & 0.833 & 299 & 16.11 & 0.257\\
OT-matched real & DINOv2 & tokenwise & 1 & 250--2.5k & 19.20 & 0.171 & 0.843 & 268 & 22.63 & 0.198\\
random same-class & DINOv2 & tokenwise & 1 & 250--2.5k & 19.62 & 0.149 & 0.858 & 278 & 24.08 & 0.200\\
OT-matched real & DINOv2 & tokenwise & 0.25 & 250--2.5k & 16.07 & 0.230 & 0.804 & 261 & 16.98 & 0.244\\
OT-matched real & DINOv2 & tokenwise & 0.1 & 250--2.5k & 15.39 & 0.252 & 0.790 & 266 & 14.94 & 0.261\\
OT-matched real & DINOv2 & tokenwise & 0.05 & 250--2.5k & 14.97 & 0.252 & 0.793 & 270 & 14.28 & 0.276\\
\quad + 2.5k pure-OT updates (5k total) & & & & & 13.27 & 0.290 & 0.781 & 263 & 13.51 & 0.299\\
OT-matched real & DINOv2 & tokenwise & 1 & 250--750 & 14.92 & 0.242 & 0.799 & 268 & 14.07 & 0.271\\
OT-matched real & DINOv2 & tokenwise & 0.25 & 250--750 & 14.99 & 0.252 & 0.782 & 268 & 13.98 & 0.287\\
\midrule
\multicolumn{5}{@{}l}{paired path: \method initialization, no alignment in post-training} & \textbf{12.89} & \textbf{0.469} & 0.788 & 269 & \textbf{12.36} & \textbf{0.502}\\
\bottomrule\end{tabular}
\caption{\textbf{All generative-path alignment configurations} (B-160 initialization, 2.5k post-training updates). Every configuration lowers recall by 0.20--0.30, for both EMA and online weights. The indented row continues the $\lambda{=}0.05$ run with 2.5k further OT updates without alignment; recall recovers only to 0.290, far below the 2.5k-update baseline (0.452). The last row is the paired-path counterpart under the same protocol: a \method-B initialization post-trained without any alignment.}
\label{tab:s-generated}
\end{table*}

\begin{table}[t]
\centering\footnotesize
\setlength{\tabcolsep}{1.6pt}
\resizebox{\linewidth}{!}{\begin{tabular}{@{}l ccccc cc@{}}\toprule
Model & Incep. & CLIP & SigLIP & ViT-MAE & DINOv2$^\dagger$ & \fdr{5} & w/o DINOv2\\\midrule
Priming-B (160) & 0.836 & 22.14 & 20.07 & 6.79 & 6.49 & 11.27 & 12.46\\
\method-B (160) & 0.780 & 22.15 & 16.79 & 5.82 & 5.47 & 10.20 & 11.38\\
\midrule
Priming-B (200) & 0.809 & 21.61 & 19.27 & 6.61 & 6.27 & 10.91 & 12.07\\
\method-B (200) & 0.772 & 20.75 & 16.03 & 5.51 & 5.25 & 9.66 & 10.76\\
\midrule
Priming-L (160) & 0.738 & 15.13 & 12.83 & 4.05 & 4.32 & 7.41 & 8.19\\
\method-L (160) & 0.730 & 13.93 & 10.44 & 3.48 & 3.57 & 6.43 & 7.14\\
\midrule
Joint-B + refine & 0.768 & 22.22 & 21.51 & 6.69 & 6.42 & 11.52 & 12.80\\
\methodJ-B + refine & 0.748 & 23.18 & 17.80 & 5.78 & 5.52 & 10.60 & 11.88\\
\bottomrule\end{tabular}}
\caption{\textbf{Normalized per-space Fr\'echet distances} of the official \fdr{5} evaluation (lower is better), averaged over all sampling seeds (three per model, six for \method-B at 160 epochs). $^\dagger$DINOv2 is our teacher, so the last column recomputes the average without it.}
\label{tab:s-components}
\end{table}

\begin{table}[t]
\centering\footnotesize
\begin{tabular}{@{}rcccc@{}}\toprule
& \multicolumn{2}{c}{Priming 40ep} & \multicolumn{2}{c}{\method 40ep}\\\cmidrule(lr){2-3}\cmidrule(l){4-5}
OT updates & \fdr{6} & Rec. & \fdr{6} & Rec.\\\midrule
1.25k & 18.11 & 0.504 & 14.61 & 0.532\\
2.5k & 16.64 & 0.538 & 13.71 & 0.566\\
5k & 15.89 & 0.563 & 13.67 & 0.591\\
10k & 14.60 & 0.586 & 12.76 & 0.607\\
15k & 14.10 & 0.583 & 12.21 & 0.606\\
20k & 13.86 & 0.597 & 11.92 & 0.619\\
22.5k & 13.71 & 0.595 & 12.00 & 0.622\\
25k & 13.60 & 0.600 & 11.58 & 0.608\\
\bottomrule\end{tabular}
\caption{\textbf{40-epoch initializations, long post-training} (online weights). After 2.5k updates, \method matches Priming after 22.5k updates.}
\label{tab:s-init40}
\end{table}

\begin{table}[t]
\centering\footnotesize
\begin{tabular}{@{}rcc@{}}\toprule
Updates & Joint (FID / \fdr{5}) & \methodJ (FID / \fdr{5})\\\midrule
50k & 3.68 / 16.71 & 4.84 / 13.96\\
60k & 3.53 / 15.53 & 4.58 / 13.28\\
70k & 3.27 / 14.70 & 4.30 / 12.67\\
80k & 3.06 / 14.12 & 3.99 / 12.12\\
90k & -- & 3.84 / 11.74\\
100k & 2.73 / 13.04 & 3.72 / 11.45\\
110k & 2.63 / 12.63 & 3.62 / 11.20\\
120k & 2.59 / 12.44 & 3.52 / 10.98\\
130k & 2.54 / 12.10 & 3.39 / 10.68\\
140k & 2.35 / 11.95 & 3.11 / 10.46\\
150k & 2.25 / 11.69 & 3.03 / 10.18\\
160k & 2.24 / 11.57 & 2.85 / 10.08\\
170k & 2.16 / 11.41 & 2.74 / 9.97\\
180k & 2.10 / 11.30 & --\\
190k & 2.08 / 11.17 & --\\
200k & 2.06 / 11.00 & 2.47 / 9.75\\
240k & 2.05 / 10.54 & 2.22 / 9.33\\
\bottomrule\end{tabular}
\caption{\textbf{Native joint models} evaluated directly from noise (EMA weights, one seed). The two runs follow the same schedule; checkpoints beyond 200k come from a continuation to 240k updates.}
\label{tab:s-native}
\end{table}

\begin{table}[t]
\centering\footnotesize
\begin{tabular}{@{}cccc@{}}\toprule
Block & \fdr{6} (+1.25k OT, online) & native FID & native \fdr{5}\\\midrule
3 & 13.06 & 6.12 & 15.75\\
5 & 12.44 & 4.84 & 13.96\\
8 & 12.87 & 4.50 & 14.63\\
12 & 13.41 & 3.90 & 15.85\\
\bottomrule\end{tabular}
\caption{\textbf{Alignment depth in \methodJ} (40 epochs). \fdr{}-type metrics prefer block 5, whereas native FID keeps improving with depth. Two-stage \method prefers block 8 (main paper, Tab.~2).}
\label{tab:s-jointdepth}
\end{table}

\section{Additional analyses}
\label{sec:s-analysis}
\subsection{Mechanism: interface and semantic routing}
\label{sec:s-mech}
\begin{figure*}[t]
\centering
\includegraphics[width=\textwidth]{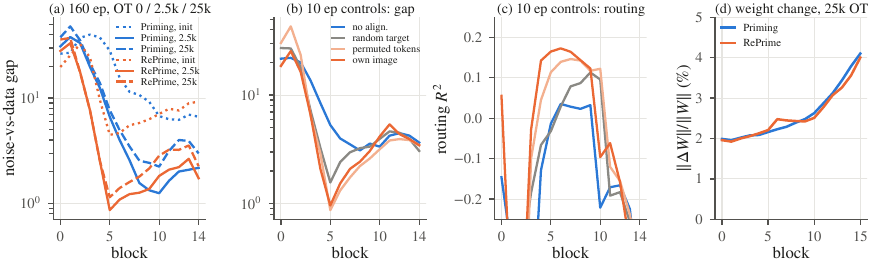}
\vspace{-12pt}
\caption{\textbf{Mechanism details} (online weights). (a) Noise-vs-data gap by block for the 160-epoch pair before post-training and after 2.5k and 25k OT updates. (b,c) Gap and semantic routing by block for the 10-epoch correspondence controls after 2.5k OT updates; all aligned arms form an interface at block 5, but only instance-paired targets route semantics through it. (d) Relative per-block weight change during 25k OT updates.}
\label{fig:s-mech}
\end{figure*}
\noindent\textbf{Definitions.} We use 1,024 ImageNet images (16 classes $\times$ 64). In \emph{data mode}, the generator receives $x_s=0.2\,x_0+0.8\,\epsilon$; in \emph{noise mode}, it receives $z\sim\mathcal N(0,I)$ with the same labels. For every block, we mean-pool token features. The \emph{noise-vs-data gap} is the Fr\'echet distance between the two feature sets divided by the trace of the data-mode covariance; the \emph{interface} is the block with the smallest gap after 2.5k OT updates, excluding the last block, which feeds the pixel output and has a small gap in every model. The \emph{spread ratio} is the trace of the within-class covariance of noise-mode features divided by that of data-mode features. For \emph{semantic routing}, we fit a ridge probe from data-mode features to the DINOv2 embedding of the clean image, apply it to noise-mode features, and report $R^2$ against the DINOv2 embedding of the image generated from that noise. The CLIP probe fits data-mode features to held-out CLIP embeddings of the clean image (\cref{sec:s-probe}). \fdr{6} values in the main paper's Fig.~5b are from online weights after 2.5k OT updates.

\noindent\textbf{Post-training is a small perturbation.} During 25k OT updates, every block moves by only 2--4\% of its norm, almost identically for both recipes (\cref{fig:s-mech}d). The structure that post-training uses must therefore come from priming.

\noindent\textbf{The interface persists through post-training.} For the 160-epoch pair (\cref{fig:s-mech}a), the gap at block 5 is 4.47 for \method and 28.35 for Priming before post-training. After 2.5k updates, \method's minimum is at block 5 (0.86), whereas Priming's lies at block 10 (1.25). After 25k updates, the minima drift to 1.14 and 2.23, respectively. Over the priming budget, the \method interface sits at block 5 with gap 4.0--4.5 from 10 epochs on, whereas the minimum of the baseline decreases only from 9.96 to 6.21 and stays at blocks 10--12.

\noindent\textbf{Correspondence controls.} \Cref{tab:s-mech} and \cref{fig:s-mech}(b,c) report the diagnostics for the four 10-epoch arms of the main paper's correspondence control. All three aligned arms form an interface at block 5, and the random-target arm has the smallest gap before post-training (2.74 vs.\ 3.74 permuted and 4.06 own image). Its features, however, do not carry instance semantics: the data-mode CLIP probe equals that of no alignment, and after 2.5k OT updates its noise inputs route almost nothing through the interface, with a low spread ratio, \ie, noise codes of a class lie close together. Permuted tokens keep the per-image token set and recover most of the routing and spread, matching their intermediate \fdr{6}. The pattern persists after 5k OT updates: at block 5, the gap is 6.43, 1.84, 1.10, and 1.14, routing is 0.03, $-0.03$, 0.08, and 0.19, and the spread ratio is 0.19, 0.26, 0.88, and 0.76 for no alignment, random targets, permuted tokens, and the own image.

\begin{table}[t]
\centering
\footnotesize
\setlength{\tabcolsep}{3pt}
\begin{tabular}{@{}l c c c c c@{}}
\toprule
Target (10 ep) & gap$_5$ init & CLIP $R^2_5$ & gap$_5$ & route$_5$ & spread$_5$\\
\midrule
no alignment & 18.10 & .360 & 5.30 & $-.01$ & .20\\
random same-class & 2.74 & .361 & 1.57 & $-.03$ & .26\\
own, permuted tokens & 3.74 & .406 & 0.87 & .11 & .85\\
own image & 4.06 & .413 & 0.96 & .17 & .76\\
\bottomrule
\end{tabular}
\caption{\textbf{Interface diagnostics for the correspondence controls} at block 5, before post-training (``init'', data-mode CLIP probe) and after 2.5k OT updates (gap, routing, spread). Online \fdr{6} after 2.5k updates: 24.24, 23.89, 21.75, 20.28; EMA \fdr{6} after 5k updates: 22.10, 21.62, 19.94, 18.10.}
\label{tab:s-mech}
\end{table}

\noindent\textbf{Generative-path alignment.} With alignment on the generative path (block 8), the gap at the aligned block is small (0.47--0.61) because the alignment pulls the class means of noise features 2--4$\times$ further from the data means while shrinking within-class spread (0.13 for random targets vs.\ 0.41 without alignment). Routing at block 5 falls to 0.10--0.16 (0.20 without alignment), and sample diversity in DINOv2 space falls from 0.376 to 0.311--0.345. A small gap alone is thus not diagnostic: across the 16 models of the main paper's Fig.~5b, its Spearman correlation with \fdr{6} is 0.00, whereas routing reaches $-0.90$ (21 of 27 within-group pairs concordant).

\noindent\textbf{Open point.} The weakest generative-path arm (OT-matched targets, $\lambda{=}0.05$) loses recall (0.252) but shows no collapse at the code level (diversity 0.372, spread 0.44, routing at its interface 0.34). Its coverage loss must therefore arise in later blocks or in the decoding of codes to pixels, which our block-pooled diagnostics do not resolve.

\subsection{Representation probes}
\label{sec:s-probe}
\begin{figure}[t]
\centering
\includegraphics[width=\linewidth]{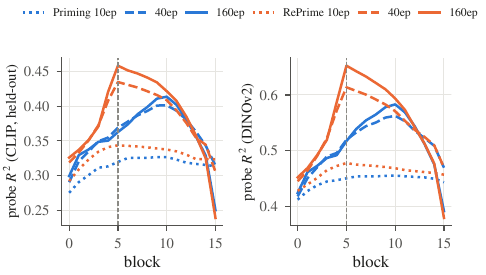}
\caption{\textbf{Linear readability of semantics} from block-wise generator features of fixed noisy real inputs ($s=0.8$), using ridge probes to the clean image's pooled CLIP (left, never used in training) and DINOv2 (right) embeddings. Validation $R^2$ on held-out images; priming checkpoints after 10, 40, and 160 epochs.}
\label{fig:s-probe}
\end{figure}
We feed the same 1,024 noisy real images (16 classes $\times$ 64, $s=0.8$, fixed noise) to priming checkpoints of both recipes after 10, 40, and 160 epochs, and fit ridge probes from mean-pooled block-$\ell$ features to the clean image's pooled DINOv2 and CLIP embeddings, with 48 training and 16 validation images per class (\cref{fig:s-probe}). \method features are more predictive at every budget and in both spaces, including CLIP, which plays no role in training: at block 5, the CLIP $R^2$ is 0.34/0.44/0.46 for \method versus 0.32/0.37/0.36 for Priming at 10/40/160 epochs. The advantage peaks at the aligned block. After 40 epochs, \method (0.435 at block 5) already exceeds the 160-epoch baseline at its best block (0.414 at block 10). This is consistent with the faster convergence of the subsequent OT stage, which operates on these features. On the same inputs, the 160-epoch reconstruction is closer in DINOv2 space to its own clean image than to its class prototype (cosine 0.81 vs.\ 0.73), \ie, the paired-path input carries instance information about the alignment target ($\mathcal A>0$ in Proposition~1).

\subsection{How selective is the OT plan?}
\label{sec:s-otplan}
Generative-path alignment with OT-matched targets assumes that the plan pairs each sample with a semantically corresponding real image. We test this at the start of post-training. For each priming checkpoint above, we generate 64 samples per class from pure noise, compute the Stage-2 Sinkhorn plan against 128 real references per class in the ConvNeXt/MAE matching space (averaged over its feature-wise problems), and measure DINOv2 cosine between each sample and (i)~its plan-weighted reference, (ii)~the reference with the largest plan mass, (iii)~a uniformly random reference of the same class, and (iv)~its nearest reference. Across all six checkpoints, (i) exceeds (iii) by at most 0.002 and (ii) exceeds (iii) by at most 0.012 (\eg, 0.221 vs.\ 0.213 at 160 epochs), whereas (iv) reaches 0.31--0.40. Seen by a semantic encoder, the plan therefore chooses targets nearly at random within the class, which explains why OT-matched and random same-class targets behave alike (main paper, Fig.~4; \cref{tab:s-generated}).

\subsection{Independent re-runs and sampling variance}
Sampling variance is small: across three sampling seeds of the same checkpoint, the standard deviation is 0.001--0.022 FID and 0.008--0.035 \fdr{5} (main paper, Tab.~1). Re-running \emph{training} changes results more, but still far less than the effect of alignment. At 160 epochs, the released Priming-B initialization and our re-trained one give \fdr{6} 9.51/9.71 and FID 1.373/1.367. Two independent \method runs give 8.61/8.47 and 1.303/1.302. The recipe gap (0.9--1.2 \fdr{6}) is four to six times the within-recipe spread.

\subsection{Teacher space}
\begin{figure}[t]
\centering
\includegraphics[width=\linewidth]{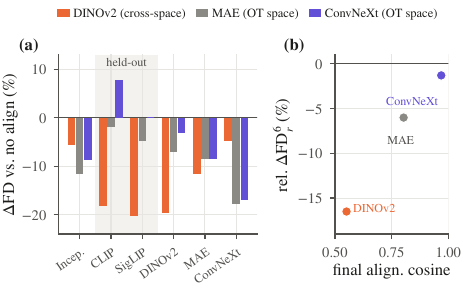}
\vspace{-14pt}
\caption{\textbf{Teacher choice} (40-epoch priming, 2.5k post-training updates, EMA). (a) Per-space change in Fr\'echet distance relative to no alignment. DINOv2 alignment improves the held-out CLIP and SigLIP spaces by 18--20\%, whereas aligning to the OT ConvNeXt space mostly improves that space and \emph{worsens} CLIP. (b) Teachers that are easy to align with yield the smallest benefit.}
\label{fig:teacher}
\end{figure}

\Cref{fig:teacher} breaks down the 40-epoch teacher comparison of the main paper by feature space. ConvNeXt alignment improves the ConvNeXt distance by 17\% but worsens CLIP by 8\%, whereas DINOv2 alignment improves the held-out CLIP and SigLIP spaces by 18--20\%; teachers that are easy to align with yield the smallest benefit (\cref{fig:teacher}b).

At full budget (160 epochs + 25k updates), aligning to the OT feature spaces does not help. ResNet-MAE alignment gives \fdr{6} 9.65 and ConvNeXt alignment 9.95 (two post-training runs from the same initialization: 9.95/9.95), versus 9.51 without alignment and 8.61 with DINOv2. Even the per-space distance of ConvNeXt alignment in its own ConvNeXt space is worse than the baseline at full budget (33.6 vs.\ 31.6), whereas DINOv2 alignment improves it (28.0).

\subsection{Alignment cosine and loss share}
At the end of 40 priming epochs, the alignment cosine is 0.55 (DINOv2), 0.80 (MAE), and 0.97 (ConvNeXt). The corresponding weighted alignment terms make up 13\%, 6\%, and 1\% of the total priming loss. On the generative path, ConvNeXt-space targets are matched almost perfectly (max-cosine 0.998) while recall collapses (\cref{tab:s-generated}). The alignment score therefore says little about the value of the alignment.

\subsection{Joint training curves}
\begin{figure*}[t]
\centering
\includegraphics[width=\textwidth]{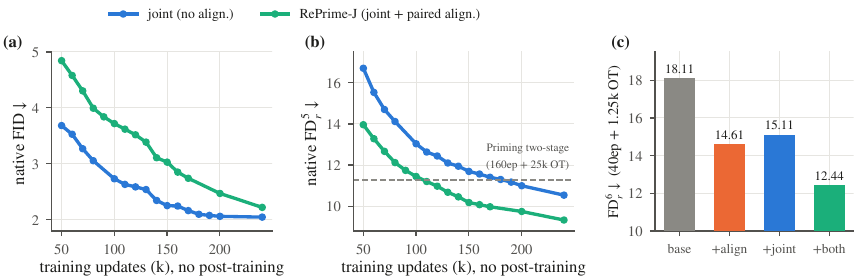}
\vspace{-14pt}
\caption{\textbf{\methodJ: joint training with paired-path alignment.} (a,b) Native FID and \fdr{5} of joint models evaluated directly from noise, without any post-training stage. Paired-path alignment trades native FID for much better \fdr{5}; the native \methodJ surpasses the \fdr{5} of the two-stage Priming-B after about 110k joint updates. (c) $2\!\times\!2$ factorial at 40 epochs (+1.25k OT updates, online \fdr{6}): alignment and joint training each help, and their combination is best.}
\label{fig:joint}
\end{figure*}

\Cref{fig:joint} shows native (no post-training) FID and \fdr{5} of joint models over training and the $2\!\times\!2$ factorial at 40 epochs (sequential vs.\ joint, with vs.\ without alignment, each followed by a 1.25k-update OT probe). The native \methodJ surpasses the \fdr{5} of the two-stage Priming-B after about 110k joint updates and reaches 9.75 after 200k updates; its native FID is higher (2.47$\to$2.22 vs.\ 2.06$\to$2.05 without alignment).

\section{Additional samples}
\label{sec:s-samples}
\begin{figure*}[p]
\centering
\includegraphics[height=0.86\textheight]{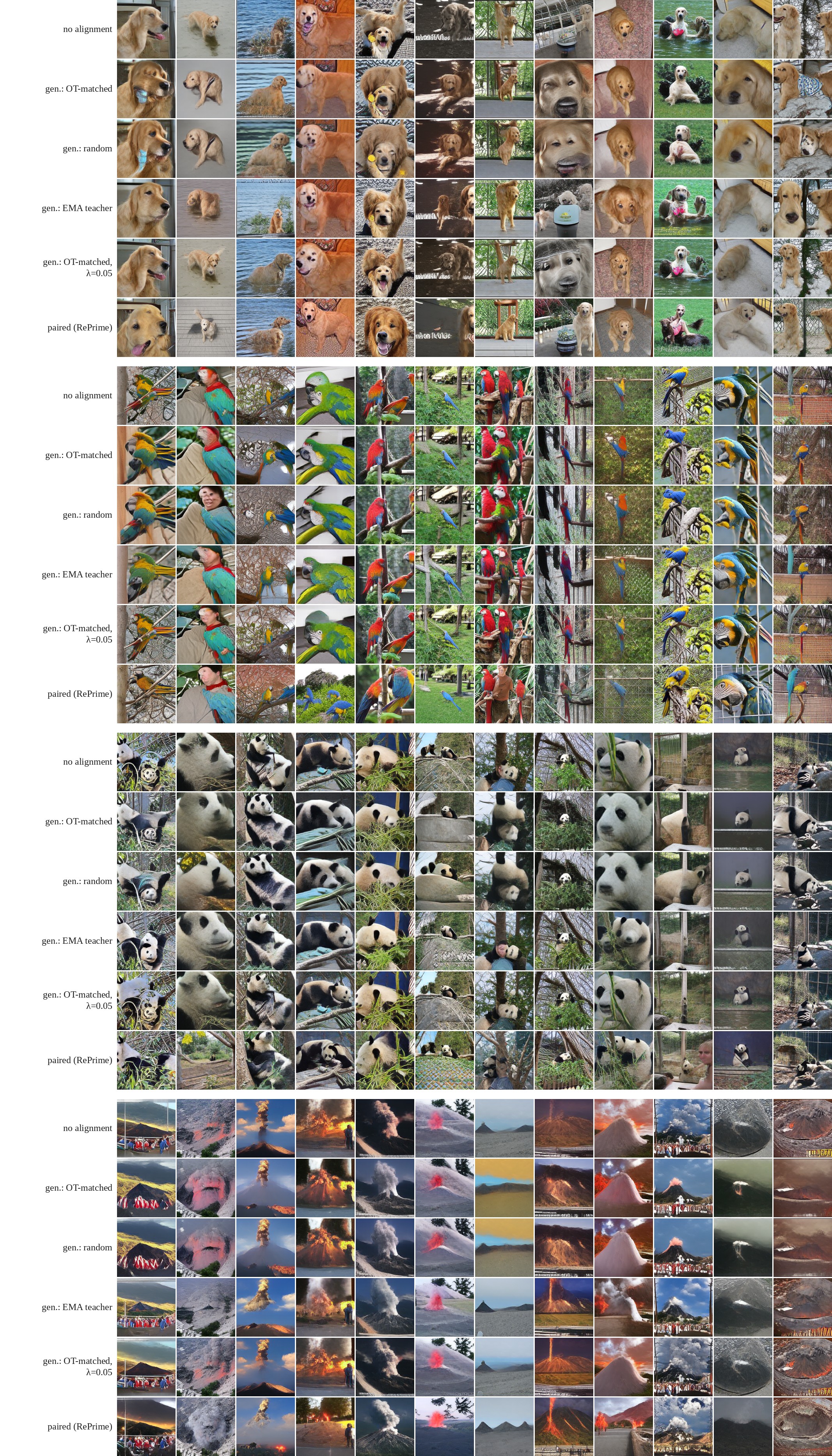}
\caption{\textbf{Prototype attraction across classes} (golden retriever, macaw, giant panda, volcano; the same 12 noise vectors in every row of a block; 2.5k post-training updates from the released Priming-B initialization, except the last row of each block, which starts from a \method initialization). In every class the OT-matched and random-target rows are nearly identical to each other, \ie, the OT-matched image acts as a class-level target. Recall and precision for these models are listed in \cref{tab:s-generated}.}
\label{fig:s-collapse}
\end{figure*}

\begin{figure*}[t]
\centering
\includegraphics[width=\textwidth]{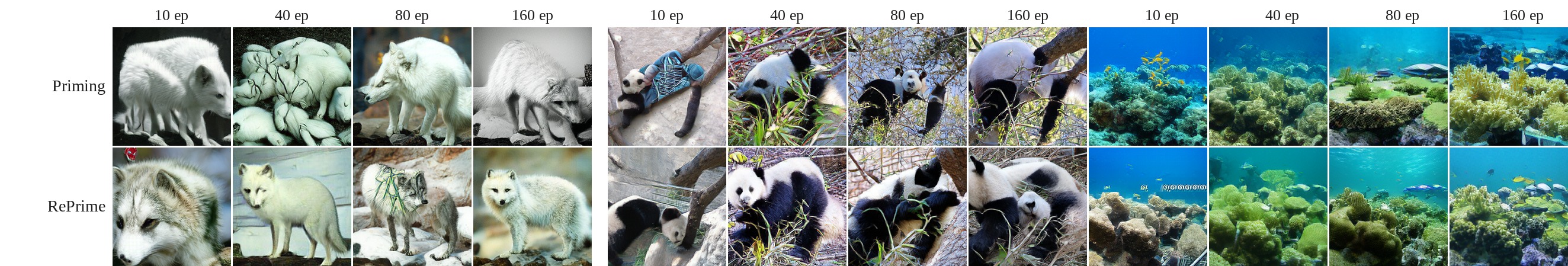}
\caption{\textbf{Same noise across priming budgets} (B, 10--160 priming epochs, each followed by 25k post-training updates; Priming top, \method bottom). Differences are largest at low budgets: at 10 and 40 epochs the Priming arctic fox is malformed, whereas \method already produces a coherent animal.}
\label{fig:s-progression}
\end{figure*}

\begin{figure*}[p]
\centering
\includegraphics[width=\textwidth]{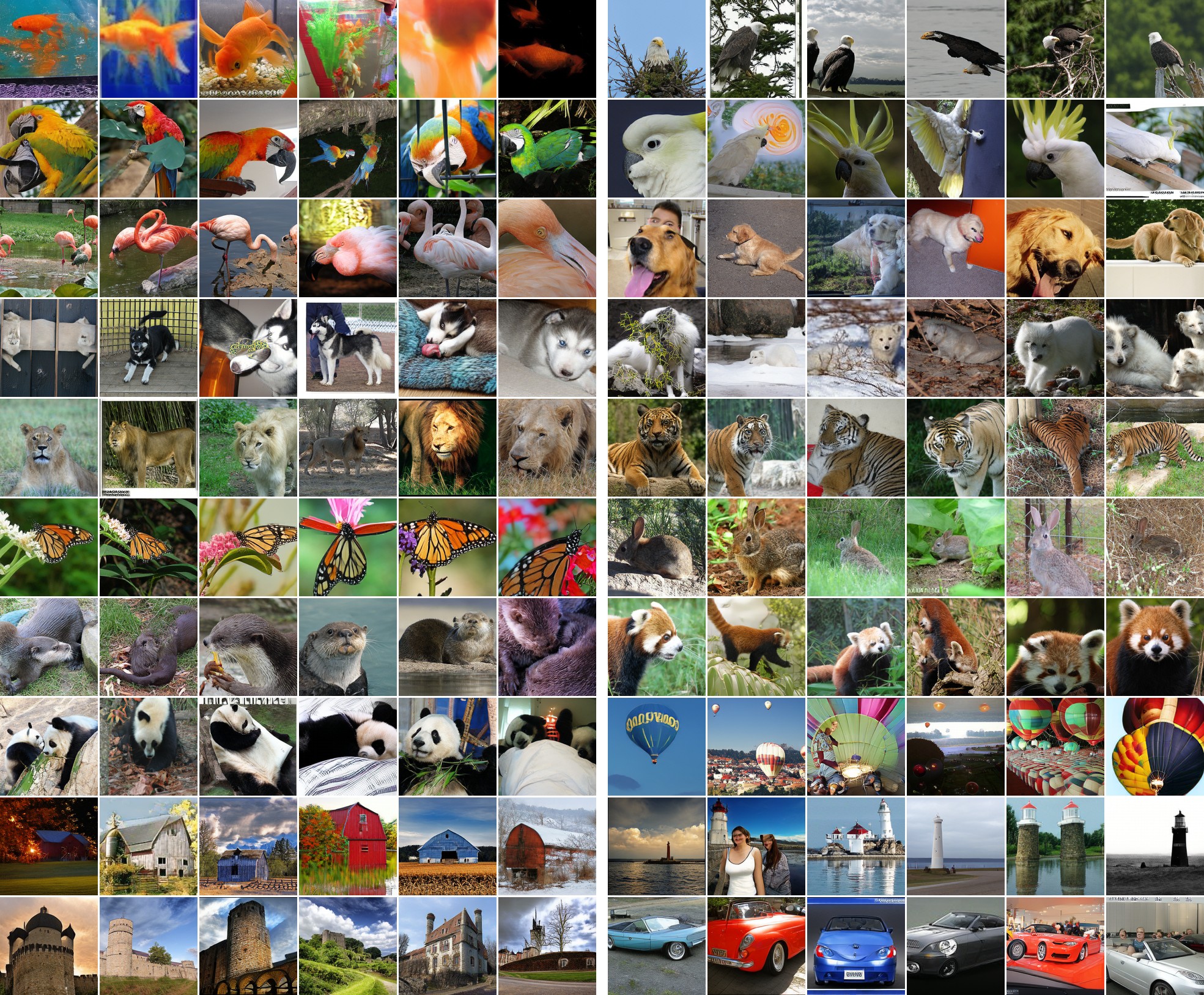}
\caption{\textbf{Uncurated \method-L samples} (FID 1.155), 1 NFE, no guidance. Two classes per row, six samples each.}
\label{fig:s-L}
\end{figure*}

\begin{figure*}[p]
\centering
\includegraphics[width=\textwidth]{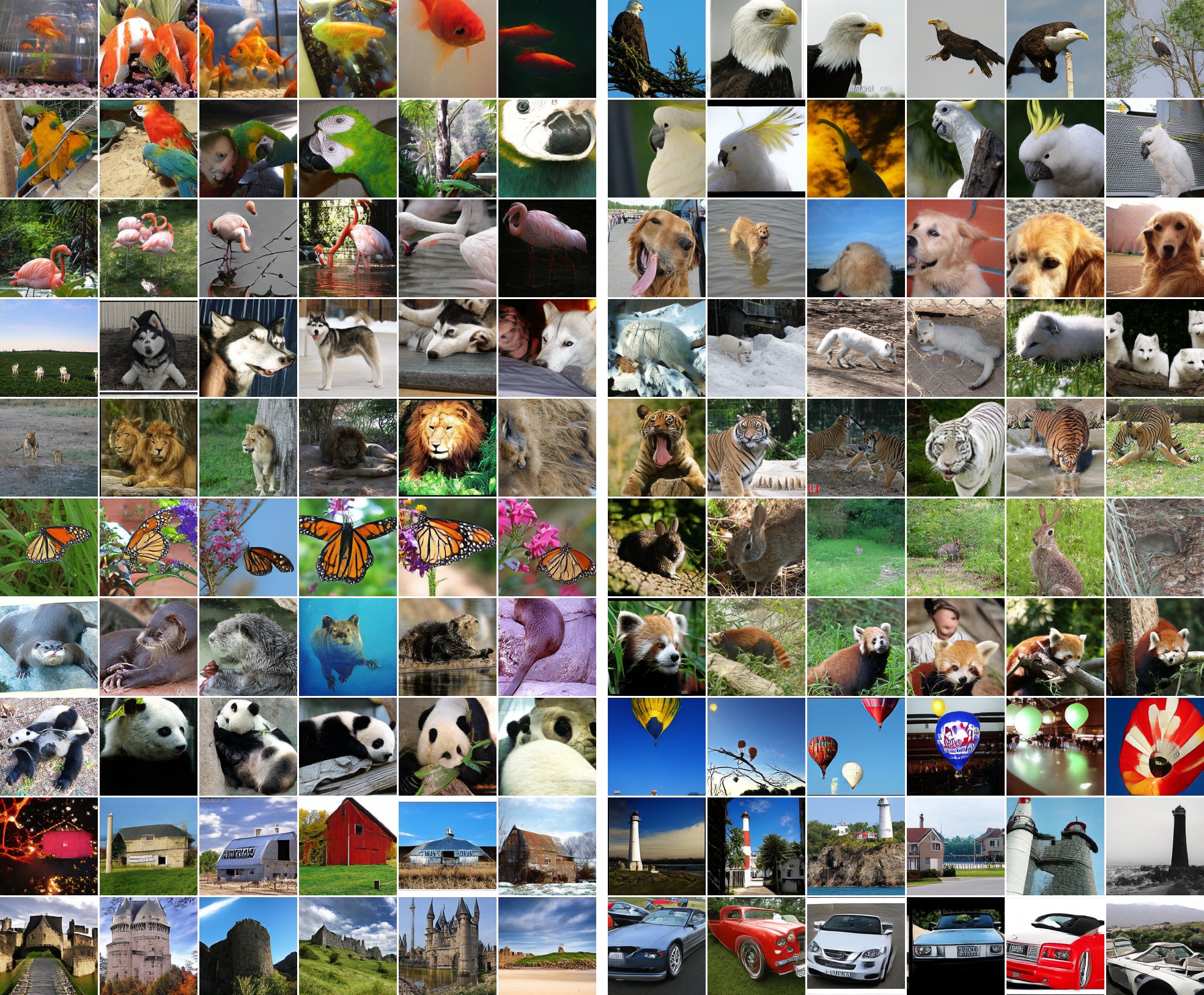}
\caption{\textbf{Uncurated \method-B samples} (block 8, 200 priming epochs, \fdr{6} 8.15).}
\label{fig:s-B}
\end{figure*}

\begin{figure*}[p]
\centering
\includegraphics[width=0.49\textwidth]{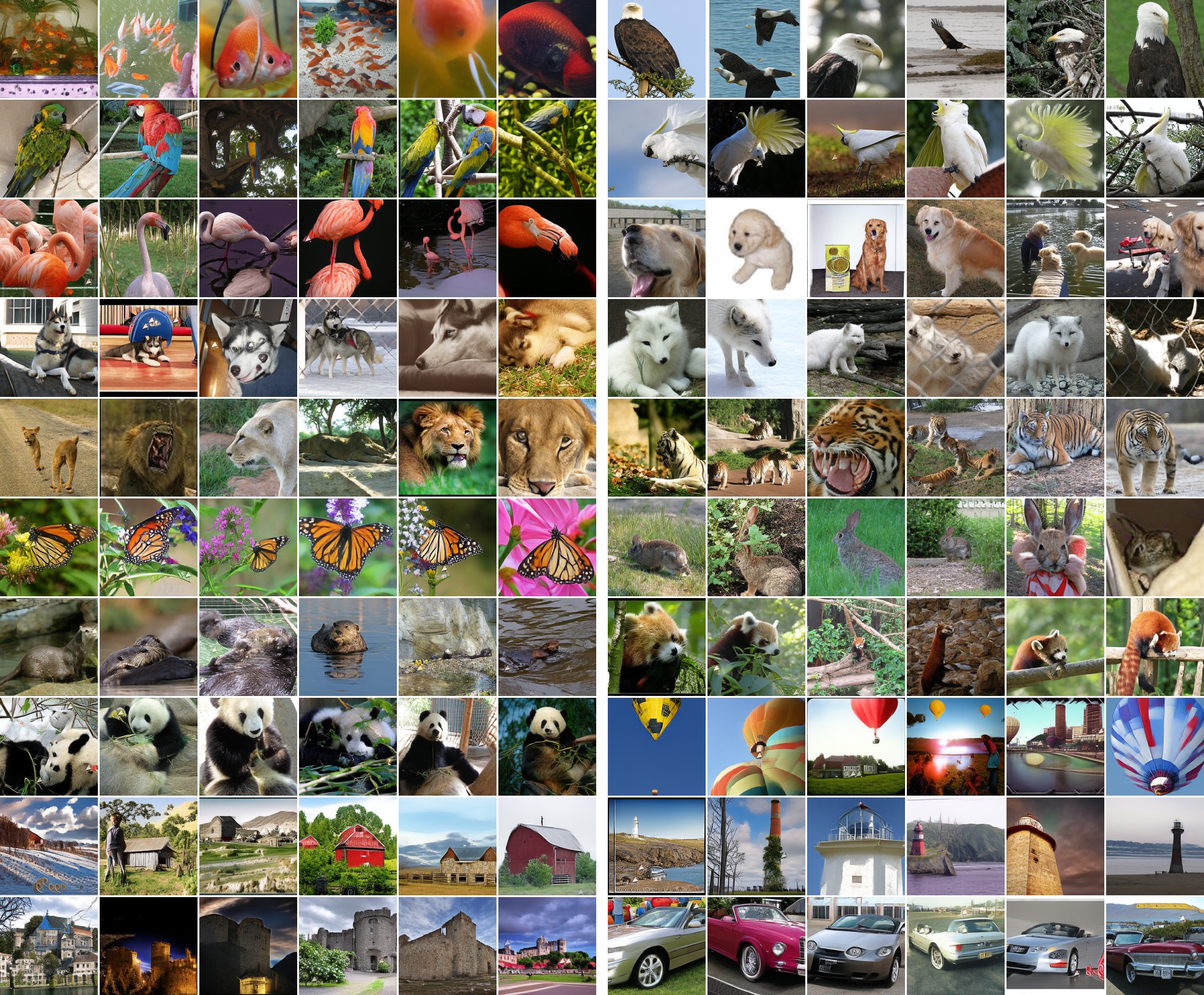}\hfill
\includegraphics[width=0.49\textwidth]{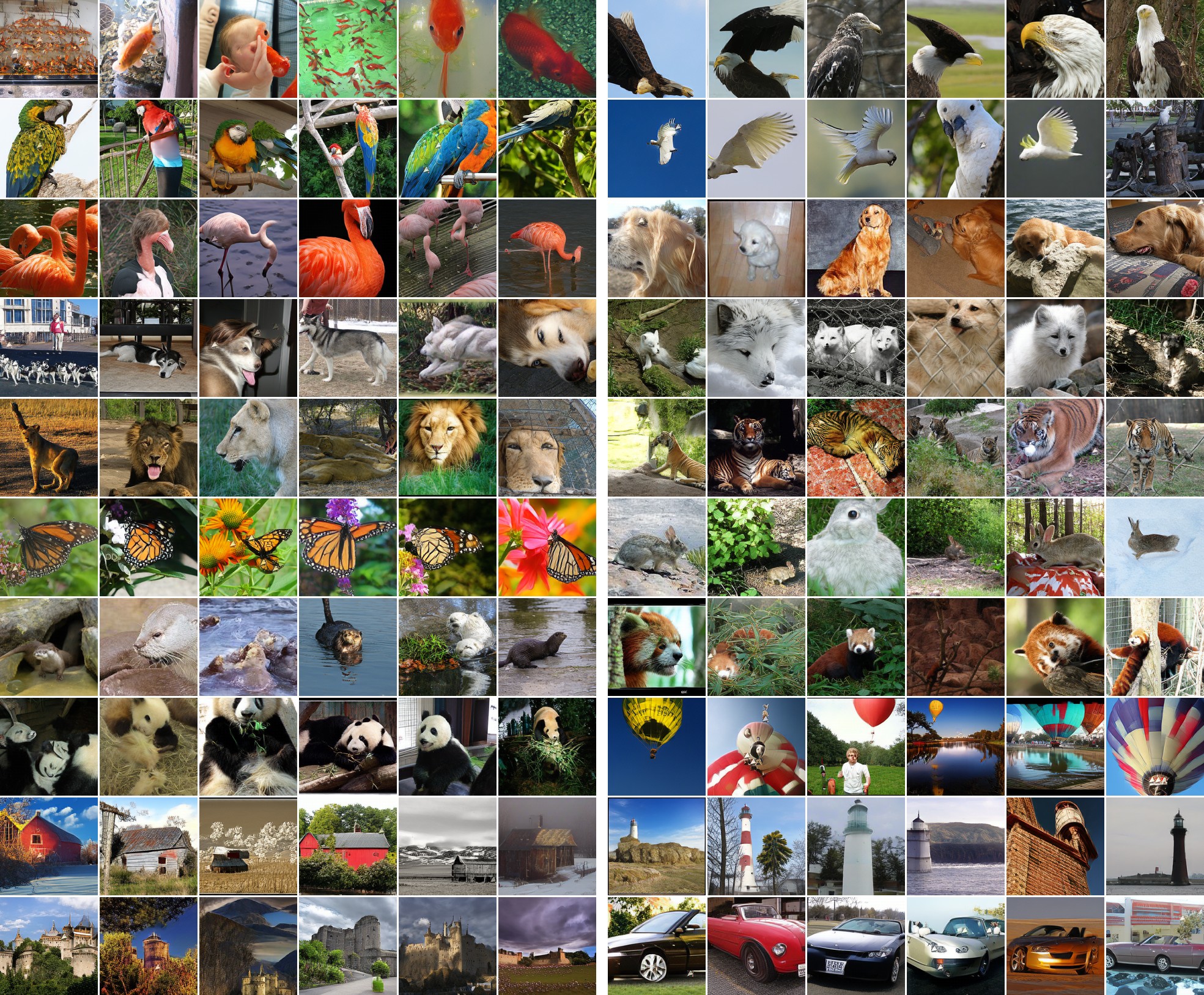}
\caption{\textbf{Uncurated \methodJ-B samples}: native model after 240k joint updates without post-training (left; FID 2.22, \fdr{5} 9.33) and after 25k refinement updates (right; FID 1.224, mean of three seeds).}
\label{fig:s-J}
\end{figure*}

\cref{fig:s-collapse} extends Fig.~4 of the main paper to four classes and twelve noise vectors, and adds the weakest generative-path setting ($\lambda{=}0.05$). \cref{fig:s-progression} follows fixed noise across priming budgets. \cref{fig:s-L,fig:s-B,fig:s-J} show uncurated samples for 20 classes with six fixed noise vectors each; no sample was selected or filtered.

\end{document}